\documentclass[10pt,twocolumn,letterpaper]{article}

\usepackage[accsupp]{axessibility}
\usepackage{multirow}
\usepackage{makecell}
\usepackage{caption}
\usepackage{arydshln}
\usepackage{subcaption}
\usepackage{array}
\usepackage[symbol]{footmisc}
\usepackage{pifont}
\usepackage{cuted}

\newcommand{\todo}[1]{{\textcolor{red}{[TODO: #1]}}}

\usepackage{iccv}      

\definecolor{iccvblue}{rgb}{0.21,0.49,0.74}
\usepackage[pagebackref,breaklinks,colorlinks,allcolors=iccvblue]{hyperref}

\def\paperID{11964} 
\def\confName{ICCV}
\def\confYear{2025}

\title{Latent Expression Generation for Referring Image Segmentation and Grounding}

\author{%
  Seonghoon Yu$^1$ \quad 
  Junbeom Hong$^1$ \quad 
  Joonseok Lee$^2$ \quad
  Jeany Son$^{3,*}$\\ [0.5ex]
  $^1$GIST \quad\quad $^2$Seoul National University \quad\quad $^3$POSTECH \\
  \tt\small \{seonghoon, joonbeom\}@gm.gist.ac.kr \; 
  joonseok@snu.ac.kr \; 
  jeany@postech.ac.kr \\
}

\begin{document}
\maketitle
\begin{abstract}



Visual grounding tasks, such as referring image segmentation (RIS) and referring expression comprehension (REC), aim to localize a target object based on a given textual description.
The target object in an image can be described in multiple ways, reflecting diverse attributes such as color, position, and more. 
However, most existing methods rely on a single textual input, which captures only a fraction of the rich information available in the visual domain.
This mismatch between rich visual details and sparse textual cues can lead to the misidentification of similar objects.
To address this, we propose a novel visual grounding framework that \textbf{leverages multiple latent expressions} generated from a single textual input by incorporating complementary visual details absent from the original description.
Specifically, we introduce {subject distributor} and {visual concept injector} modules to embed both \textbf{shared-subject and distinct-attributes} concepts into the latent representations, thereby capturing unique and target-specific visual cues.
We also propose a positive-margin contrastive learning strategy to align all latent expressions with the  original text while preserving subtle variations.
Experimental results show that our method not only outperforms state-of-the-art RIS and REC approaches on multiple benchmarks but also achieves outstanding performance on the generalized referring expression segmentation (GRES) benchmark.\footnotetext{\hspace{-2em}$^*$ indicates the corresponding author and the work was done at GIST.}

\end{abstract}    
\section{Introduction}
\label{sec:intro}


Visual grounding (VG) tasks are one of the fundamental vision-language tasks that aim to locate visual regions described by a given textual expression.
For example, referring image segmentation (RIS) predicts a segmentation mask, while referring expression comprehension (REC) detects a bounding box from the input text.
Unlike conventional segmentation and detection tasks that rely on pre-defined categories, visual grounding tasks allow free-form, open-vocabulary, and arbitrary-length text given by users to locate interested regions.
The language-based nature of this task enables its application in diverse fields, such as image editing~\cite{image_editing} and robotic manipulation~\cite{robot_manipulation}.

\begin{figure}[t]
    \centering
    \includegraphics[width=1.0\linewidth]{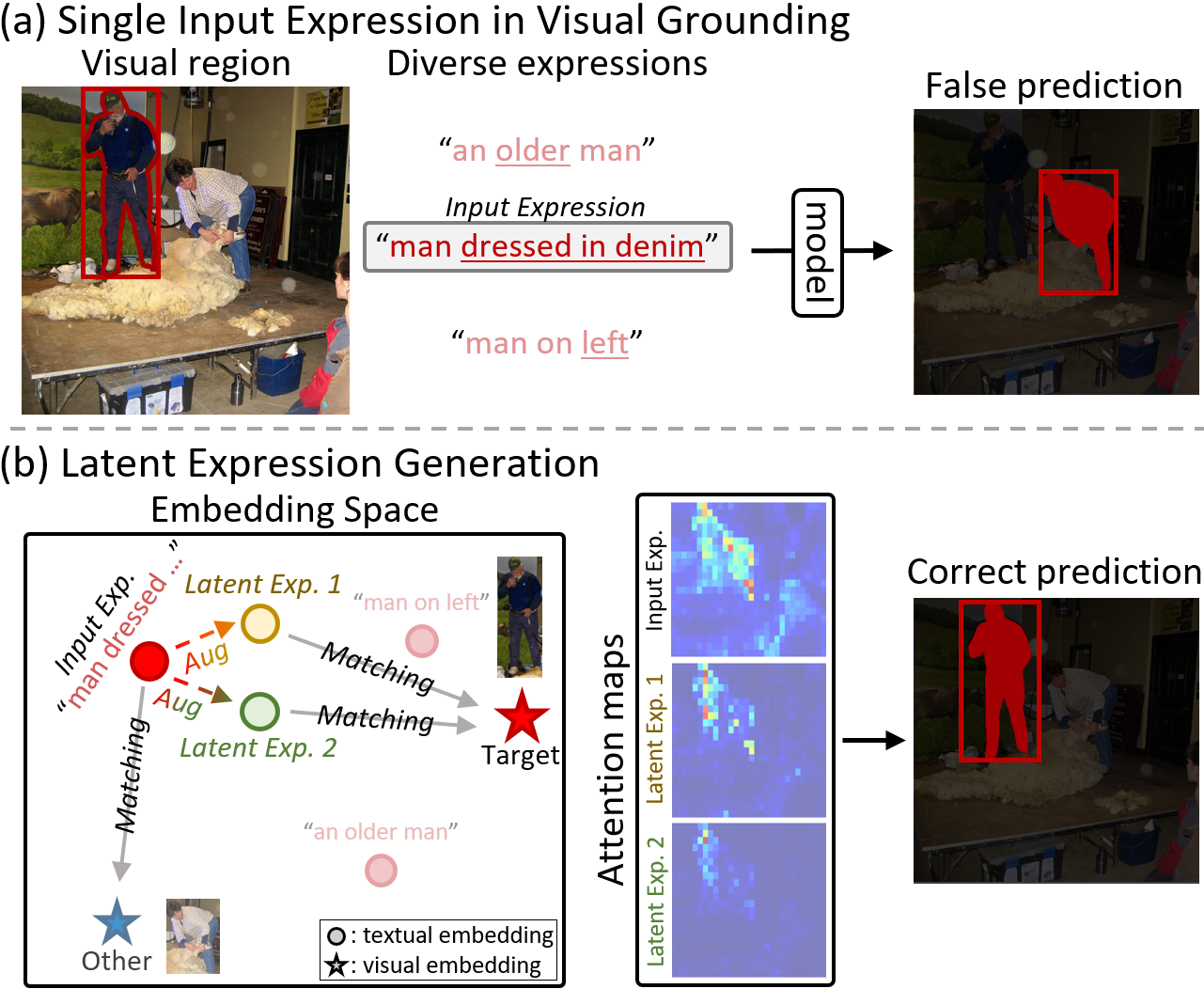}
    \vspace{-0.5cm}
    \caption{The motivation for latent expression generation: (a) visual regions can be expressed in various ways (\eg, ``\textit{an older man}", ``\textit{man dressed in denim}", ``\textit{man on left}", \etc), yet visual grounding tasks (RIS and REC) typically rely on a single, brief description (\eg, ``\textit{man dressed in demin}"), leading a model to incorrect predictions; (b) to overcome this, we augment the original text into multiple latent expressions in the latent space, each capturing different visual details as visualized in attention maps.} 
    \vspace{-0.5cm}
    %

    \label{fig:intro} 
\end{figure}

Visual regions in an image can be described in many ways using diverse textual expressions, since the visual contents cover various aspects such as color, texture, spatial arrangement, and contextual relationships between objects.
However, a single expression provided in RIS and REC for prediction often consists of a few words, capturing only a subset of the rich visual details present in the target regions (See Fig.~\ref{fig:intro}a).
This limitation leads to a semantic gap between the input description and the detailed visual information of the target region, causing the model's prediction to depend heavily on the specific textual cues given.

For example, although the target in Fig.~\ref{fig:intro}a can be described with various descriptions, the model is provided with only a single expression (\eg, \textit{``man dressed in demin''}) to predict the target mask or bounding box.
Such a single and sparse expression can lead to confusion when similar objects are present, leading to incorrect predictions (\eg, selecting a woman wearing jeans instead of the intended target).
Our work stems from the idea that generating multiple diverse latent expressions from the input text—each capturing complementary visual details while preserving the shared subject—and leveraging them collectively can overcome this challenge.
As illustrated in Fig.~\ref{fig:intro}a, the expressions ``\textit{man dressed in denim}", ``\textit{an older man}" and ``\textit{man on left}" all refer to the same subject (``\textit{man}"), yet highlight different attributes such as ``\textit{denim}", ``\textit{older}" and ``\textit{left}".
Similar to this, we focus on injecting distinct visual cues into each generated latent expression that refers to the same target.
Consequently, the attention maps (Fig.~\ref{fig:intro}b) derived from our generated latent expressions highlight different regions of the target object, thereby enriching the overall representation.

Building on this idea, we propose a novel visual grounding framework that augments a single textual input into multiple latent expressions, enabling the model to capture additional semantics not covered by the original text.
To achieve this, we first initialize the latent expressions by augmenting the input text, randomly dropping input text tokens and varying the expression length to capture different attributes.
Next, to infuse novel target visual cues into these generated latent expressions, we introduce two modules—\textit{subject distributor} and \textit{visual concept injector}—based on the principles of \textbf{shared-subject and distinct-attributes}.
Specifically, the \textit{subject distributor} ensures that all latent expressions share the same subject, while the \textit{visual concept injector} injects distinct visual details of the target into the attribute tokens.
We also introduce a positive-margin contrastive learning strategy, which aligns these latent expressions with the original text while maintaining subtle differences, thereby fostering a diverse set of expressions that retain the input context.

By leveraging these generated latent expressions for mask prediction, our framework effectively bridges the semantic gap between rich visual details in the referenced regions and limited details in the textual input.
Our extensive experiments validate that our approach not only achieves superior performance on referring image segmentation (RIS) benchmarks but also shows outstanding performance on referring expression comprehension (REC) benchmarks without requiring any task-specific decoder.
Moreover, although our method is not specifically designed for generalized referring expression segmentation (GRES), it outperforms state-of-the-art methods on GRES benchmarks with only minimal modifications for no-target cases.

Our contributions can be summarized as follows:
\begin{itemize}
    \item We propose a novel visual grounding framework that augments a single input description into multiple latent expressions, each preserving the same subject while capturing diverse, complementary visual attributes.
    \item The proposed subject distributor and visual concept injector modules allow the model to achieve the shared-subject and distinct-attributes characteristics, enabling each expression to highlight unique visual details of the target. 
    \item We introduce a positive-margin contrastive loss that aligns the generated latent expressions with the original text while avoiding identical representations by allowing the margin of differences between them.
    \item Our experiments demonstrate that the proposed method outperforms state-of-the-art RIS and REC methods across various benchmarks, and it achieves remarkable performance on the GRES benchmark as well.
    \vspace{-0.2cm}


\end{itemize}

\section{Related Work}
\label{sec:related}

\paragraph{Visual Grounding.}
Visual grounding (VG) aims to locate the visual regions referred by a given textual description via a segmentation mask in referring image segmentation (RIS) or a bounding box in referring expression comprehension (REC).
\textbf{For RIS}~\cite{nemo, pseudo-ris}, early works~\cite{related_ris_cnn1, related_ris_cnn2} rely on convolutional layers to extract visual features and RNN or LSTM to capture textual features, followed by processing the concatenated features for mask generation. 
The advanced methods~\cite{late_2_restr, joint_2_coupalign, joint_3_crossvlt, 80_classes} utilize separate transformer-based encoders for multi-modal feature extraction and fuse them via cross-attention modules.
Recent works~\cite{late_3_cris, oneref, sharedris} have adapted pre-trained vision-language models, such as CLIP~\cite{clip} and BEiT-3~\cite{beit3}, as encoders to enhance vision-language alignment.
\textbf{For REC}~\cite{clipvg}, similar to RIS, early works explored region-based approaches~\cite{related_rec_cnn1, mattnet}, later adapting Transformer models for cross-modal alignment~\cite{mdetr, transvg++, grounding_dino, eevg}, with recent methods leveraging pre-trained vision-language models for improved grounding~\cite{hivg, dynamicmdetr, oneref}. 
{Current visual grounding (RIS and REC) methods rely on a single, short expression that lacks textual cues for target localization, whereas our approach exploits extra cues outside of the original input by producing latent expressions with distinct visual details.}
\vspace{-0.5cm}

\begin{figure*}[t]
    \centering
      \includegraphics[width=0.96\textwidth]{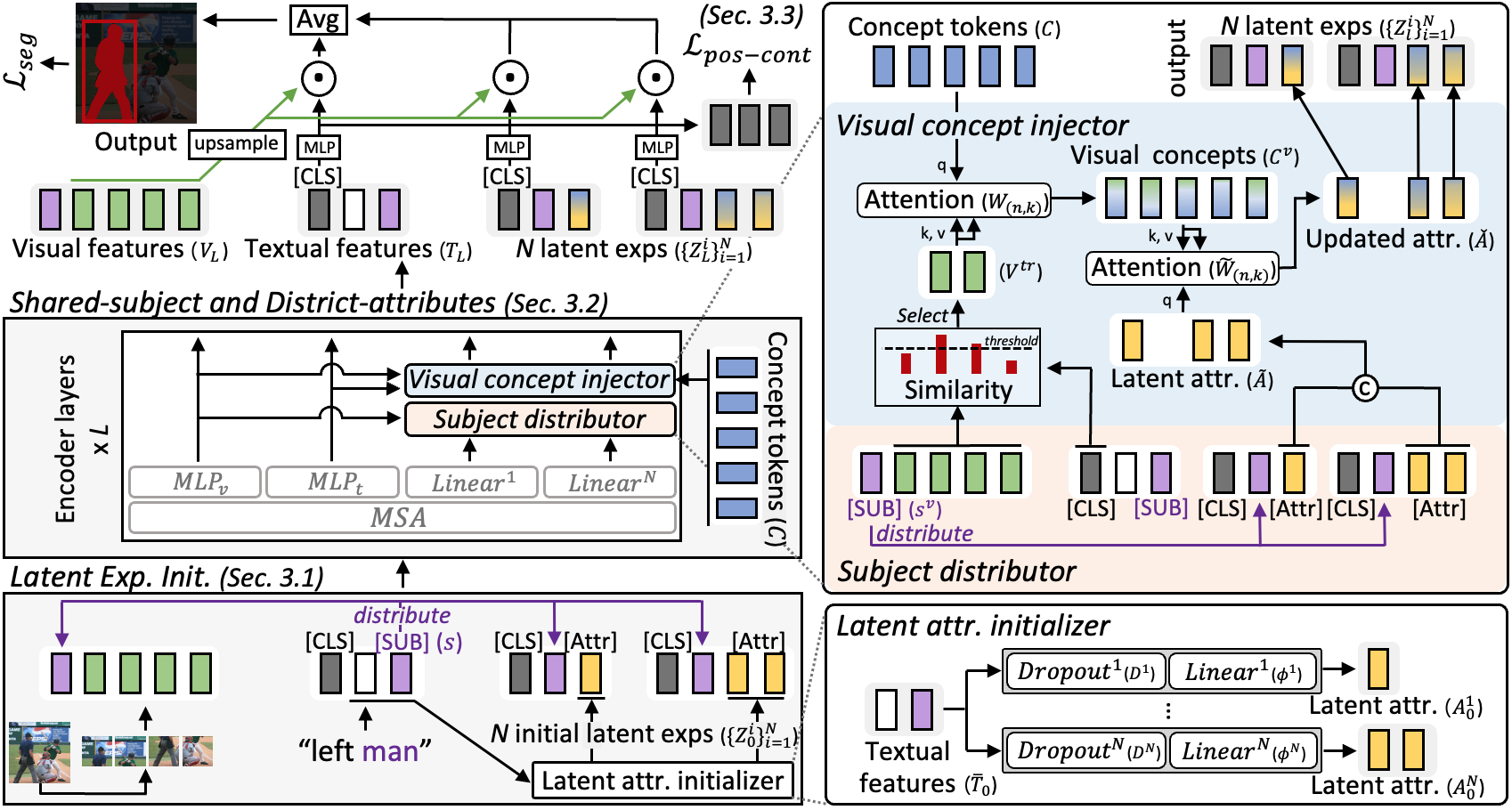}
      \vspace{-0.2cm}
    \caption{The illustration of our Latent-VG. Given an input image and text, we first generate initial multiple latent expressions (Sec.~\ref{subsec:latent generator}). These latent expressions are then concatenated with visual and textual embeddings and passed through the enoder. Within each layer, all latent expressions consistently point to the same subject, while featuring distinct visual concepts (Sec.~\ref{subsec:latent encoder}). Finally, our learning objective (Sec.~\ref{subsec:learning objective}) ensures that the latent expressions {refer to the specified target in the input text, while allowing for variations in representation.}}
    \label{fig:framwork} 
    \vspace{-0.4cm}
\end{figure*}

\paragraph{Text Augmentation.}
Text augmentation has been extensively explored to improve the robustness and generalization of language models~\cite{llava, vilbert}.
The traditional approaches~\cite{eda, back_translation}, such as text replacement~\cite{eda} and back-translation~\cite{back_translation}, aim to generate diverse yet semantically consistent variations of textual data.
SimCSE~\cite{simcse}, instead of explicitly transforming text, applies dropout to make variations in the latent space.
E-Mix~\cite{emix} also performs text augmentation in the latent space by interpolating hidden representations.
Similarly, our method augments input text in the latent space within the encoder, but has a key difference: we encourage a shared subject and distinct attributes across multiple latent expressions so that each one represents the same target with diverse visual attributes.


\paragraph{Contrastive Learning.}
Contrastive learning~\cite{rankn, simclr} has been widely employed in both self-supervised and supervised learning to learn discriminative representations by aligning positive samples while pushing negatives apart. 
The objective is typically formulated using contrastive loss functions such as InfoNCE~\cite{cpc} or Triplet~\cite{triplet} losses. 
These methods have demonstrated effectiveness in various tasks, including image representation learning~\cite{simclr}, text embeddings~\cite{clear}, and multimodal learning~\cite{clip}.
In the scope of visual grounding, CRIS~\cite{late_3_cris} and CGFormer~\cite{late_9_cgformer} propose pixel-level contrastive loss for enhanced fine-grained vision-language alignments. 
The traditional contrastive learning methods aim to maximize the alignment of positive pairs,
while the proposed positive-margin contrastive loss allows a margin of variations within positives to avoid identical representations and diversify positives.


\section{Method}
\label{sec:method}
In this section, we present Latent-VG, the novel framework for visual grounding that expands a single input expression into multiple latent expressions of varying lengths—each enriched with additional attribute cues—and utilizes these expressions for mask prediction.
The primary goal of Latent-VG is to generate distinct latent expressions that capture diverse concepts beyond those in the original input text, while referring to the same target specified by the input text.

As illustrated in Fig~\ref{fig:framwork}, our framework consists of three key components: 1) the \textit{latent expression initialization} (Sec.~\ref{subsec:latent generator}), 2) the \textit{shared-subject and distinct-attributes} (Sec.~\ref{subsec:latent encoder}), where we detail the subject distributor and visual concept injector modules, and 3) the \textit{learning objective} (Sec.~\ref{subsec:learning objective}), which includes our proposed positive-margin contrastive learning.




\subsection{Latent Expression Initialization}
\label{subsec:latent generator}
To augment latent expressions from a single textual description, we first initialize them using the token embeddings of the original text. 
Specifically, we diversify these latent expressions using our \textit{latent attribute initializer}, which randomly drops semantic tokens to reduce dependency on the original text and adjusts the length of the generated expressions to convey distinct semantic nuances.
Next, to ensure that the target subject is accurately captured, we automatically select a `subject token' from the input text tokens and distribute it to both the visual and latent textual {domain}.
The detailed process is shown in Fig.~\ref{fig:framwork} and described below.
\vspace{-0.2cm}

\paragraph{Visual and Textual Embeddings.}
To employ an input image and a text description in our framework, we first embed the input image $\mathcal{V} \in \mathbb{R}^{h \times w \times 3}$ into visual patch embeddings $\mathbf{V}_0 = [\mathbf{v}_{[cls]}, \mathbf{v}_1, ..., \mathbf{v}_n] \in \mathbb{R}^{(n+1) \times d}$, and the input text $\mathcal{T}$ with $m$ words into textual token embeddings $\mathbf{T}_0 = [\mathbf{t}_{[cls]}, \mathbf{t}_1, ..., \mathbf{t}_m] \in \mathbb{R}^{(m+1) \times d}$. Here, $n = hw/p^2$ denotes the number of patches (with $p$ as the patch size), $d$ represents the channel dimension, $\mathbf{v}_{[cls]}$ is the visual class token, and $\mathbf{t}_{[cls]}$ is the textual class token.

\vspace{-0.3cm}
\paragraph{Multiple Latent Expression Generation.} 
To generate initial $N$ latent expressions $\{\mathbf{Z}^i_{0}\}_{i=1}^N$, we first produce latent attribute tokens ${\mathbf{A}}^i_0$ from the textual token embeddings without a class token $\bar{\mathbf{T}}_0 = \left[\mathbf{t}_1, ..., \mathbf{t}_m \right]  \in \mathbb{R}^{m \times d}$ via our \textit{latent attribute initializer}.
For each $i$-th latent expression, we apply a length transform layer $\phi^{i} \in \mathbb{R}^{k^i \times m}$ to convert the number of tokens from $m$ to $k^i$, as follows:
\vspace{-0.15cm}
\begin{equation}
    {\mathbf{A}}^i_{0} = \left[ \mathbf{a}^i_{0,1}, ...,\mathbf{a}^i_{0,k^i} \right] = \phi^i (\mathbf{D}^{i} \odot \bar{\mathbf{T}}_0) \in \mathbb{R}^{k^i \times d},
    \vspace{-0.1cm}
\end{equation}
where $\mathbf{D}^i \in \{ 0,1 \}^{m \times d}$ is a dropout mask with a probability $p^i$ for randomly dropping semantic tokens during training, $k^i$ is the pre-defined token length of $i$-th generated latent expression, and $\odot$ is a Hadamard product.
In the next section, these attribute tokens will be utilized to capture diverse visual cues of the target.
To ensure that the attributes decorate a specific subject in an input text, we first prepend a subject token $\mathbf{s}$, which is selected from $\bar{\mathbf{T}}_0$ using a linear layer followed by a straight-through Gumbel-softmax~\cite{gumble} operation.
Next, we prepend a latent class token $\mathbf{z}^i_{[cls]}$.
The $i$-th initial latent expression $\mathbf{Z}^{i}_{0}$ is then defined as follows:
\vspace{-0.2cm}
\begin{align}
    \mathbf{Z}^{i}_{0} &= \left[ \mathbf{z}^i_{[cls]}, \mathbf{s}, {\mathbf{A}}^i_{0} \right] \in \mathbb{R}^{(k^i+2) \times d}.
    \vspace{-0.1cm}
\end{align}
Each $\mathbf{Z}^{i}_{0}$ represents a unique initial latent expression produced using {individual latent attribute tokens}, enabling targets to be depicted with different token lengths {and nuances}.

We also replace a visual class token $\mathbf{v}_{[cls]}$ in $\mathbf{V}_0$ with a subject token $\mathbf{s}$, denoted as $\mathbf{s}^{\mathcal{V}}$:  $\mathbf{V}_0 \leftarrow [ \mathbf{s}^{\mathcal{V}}, \mathbf{v}_1, ..., \mathbf{v}_n ] \in \mathbb{R}^{(n+1) \times d}$.
This visual subject token $\mathbf{s}^{\mathcal{V}}$ in the visual domain will be distributed to all latent expressions in the next section to ensure a consistent subject reference.

\subsection{Shared-Subject and Distinct-Attributes}
\label{subsec:latent encoder}
In this section, we introduce two modules, 1) \textit{subject distributor} and 2) \textit{visual concept injector}, designed to enforce a shared-subject and distinct-attributes principle in all latent expressions across all $L$ encoder layers.
The \textit{subject distributor} ensures that the target subject is consistently referenced, while the \textit{visual concept injector} introduces diverse visual cues into the latent attribute tokens. 
Together, these modules enable our model to generate varied textual representations for the same target.
Our method is inspired by the example in Fig.~\ref{fig:intro}a, where the target can be described with diverse expressions, all referring to the same subject, ``\textit{man}," but highlighting different attributes (``\textit{left}", ``\textit{denim}," and ``\textit{older}").
We believe that by collectively considering these attributes, the target can be identified more precisely. 
However, since the input text provides only sparse attributes, we extract additional attributes from the visual domain to refine the representation further.

As in Fig~\ref{fig:framwork}, each encoder layer starts with the shared self-attention of BEiT-3~\cite{beit3} to encode three types of embeddings (\ie, visual patches $\mathbf{V}$, textual tokens $\mathbf{T}$, and latent expressions $\{\mathbf{Z}^i\}_{i=1}^{N}$).
Specifically, for each layer $l$, we concatenate all embeddings and process them through a multi-head self-attention layer, \ie, $[ \hat{\mathbf{V}}_l, \hat{\mathbf{T}}_l, \hat{\mathbf{Z}}_l^{1}, ..., \hat{\mathbf{Z}}_l^{N} ] = \text{MSA}([ \mathbf{V}_{l-1}, \mathbf{T}_{l-1}, \mathbf{Z}_{l-1}^{1}, ..., \mathbf{Z}_{l-1}^{N} ])$.
Then we use separate MoEs for each input embedding:
$\mathbf{V} = \text{MLP}_v(\hat{\mathbf{V}})$, $\mathbf{T} = \text{MLP}_t(\hat{\mathbf{T}})$, ${\mathbf{Z}}^{i} = [ {\mathbf{z}}^i_{[cls]}, {\mathbf{s}}^i, {\mathbf{A}}^i  ] = \text{Linear}^i(\hat{\mathbf{Z}}^{i}), \forall i$.
Note that we omit the layer index $l$ for simplicity here and in the following equations.

\vspace{-0.2cm}
\paragraph{Subject Distributor.} 
After passing each shared self-attention layer, the subject tokens $\{{\mathbf{s}}^{i}\}_{i=1}^{N}$ across all latent expressions $\{{\mathbf{Z}}^i\}_{i=1}^{N}$ may no longer represent the same subject or may have arbitrary semantics.
To address this, we re-align these tokens by exchanging the subject tokens $\{{\mathbf{s}}^{i}\}_{i=1}^{N}$ in $\{{\mathbf{Z}}^i\}_{i=1}^{N}$ into the visual subject token $\mathbf{s}^{\mathcal{V}}$ in ${\mathbf{v}}$:
\vspace{-0.1cm}
\begin{equation}
    {\mathbf{Z}}^i \leftarrow \left[{\mathbf{z}}^i_{[cls]}, \mathbf{s}^{\mathcal{V}}, {\mathbf{A}}^i \right] \in \mathbb{R}^{(k^i+2) \times d}. 
    \vspace{-0.1cm}
\end{equation}
This ensures that all $\{{\mathbf{Z}}^i\}_{i=1}^{N}$ consistently share the same visual subject throughout the subsequent layers.

\vspace{-0.2cm}
\paragraph{Visual Concept Injector.}
To capture distinct attribute details in the generated latent expression, we inject unique visual concepts into the attribute tokens ${\mathbf{A}}^i = [ \mathbf{a}_1^i, \dots, \mathbf{a}_{k^i}^i ]$ at each layer.
First, we initialize $N_c$ concept tokens $\mathbf{C} \in \mathbb{R}^{N_c \times d}$ using the orthogonal initialization~\cite{orthogonal_init} to ensure their distinctiveness. 
We then choose target-related patches $\mathbf{V}^{tr}$ from the visual patches $\bar{\mathbf{V}} = [ \mathbf{v}_1, \dots, \mathbf{v}_n ]$ (which do not include a class token) by thresholding their similarities with the textual class token $\mathbf{t}_{[cls]}$, as follows: 
\vspace{-0.1cm}
\begin{equation}
    {\mathbf{V}}^{tr} = \{  {\mathbf{v}}_i |  \mathbf{v}_i \mathbf{t}_{[cls]}^\top \geq \kappa,  i \in \{1,...,n\}  \} \in \mathbb{R}^{N_{tr} \times d},
\end{equation}
where $\kappa$ is a threshold, and $N_{tr}$ is the number of patches selected as target-related patches.  
By setting the threshold $\kappa$ to the mean value of similarities, a large portion of visual patches are selected as target-related patches, thus (1) reducing reliance on the input text and (2) including contextual patches beyond target regions. 
From these target-related patches ${\mathbf{V}}^{tr}$, we retrieve visual concepts by operating a weighted-mean (\ie, attention) with the concept tokens $\mathbf{C}$, resulting in visual concept tokens $\mathbf{C}^{\mathcal{V}}$:
\vspace{-0.1cm}
\begin{gather}
        \mathbf{C}^{\mathcal{V}} = \mathbf{W}{\mathbf{V}^{tr}} \in \mathbb{R}^{N_c \times d}, \\
    \text{where~~}\mathbf{W}_{(n,k)}  = \frac{e^{M_{n,k}}}{\sum_{k=1}^{N_{tr}} e^{M_{n,k}}},  ~~M =  \mathbf{C}({\mathbf{V}^{tr}})^\top. \nonumber
    \vspace{-0.2cm}
\end{gather}

We then inject these visual concept tokens $\mathbf{C}^{\mathcal{V}}$ into the concatenated latent attribute tokens $\tilde{\mathbf{A}}= [ {\mathbf{A}}^1, ..., {\mathbf{A}}^N ] \in \mathbb{R}^{N_a \times d}$ (where $N_a$ is a total number of attributes across all latent expressions $\{\mathbf{Z}^{i}\}_{i=1}^{N}$) via a weighted-mean, where we normalize the attention score over the attribute tokens:
\vspace{-0.1cm}
\begin{align}
    & ~~~~~~~~~~~~~~~{\check{\mathbf{A}}} = \tilde{\mathbf{W}} \mathbf{C}^{\mathcal{V}} \in \mathbb{R}^{N_a \times d},\\
    &\text{where} ~ \tilde{\mathbf{W}}_{(n, k)} = \frac{e^{\tilde{M}_{n,k}}}{\sum_{n=1}^{N_{a}} e^{\tilde{M}_{n,k}}}, ~~\tilde{M} = \tilde{\mathbf{A}} ({\mathbf{C}^{\mathcal{V}}})^\top. \nonumber
    \vspace{-0.3cm}
\end{align}
By normalizing the attention map $\tilde{M}$ across attribute tokens, we encourage each attribute token to compete for and bind a distinct set of visual concept tokens, similar to the slot-attention~\cite{slotattention}, leading attribute tokens to contain distinct visual concepts.
This new concatenated attribute tokens $\check{\mathbf{A}}$ are then split back to comprise the latent expression $\mathbf{Z}^i$:
\vspace{-0.1cm}
\begin{equation}
    {\mathbf{Z}}^i \leftarrow \left[{\mathbf{z}}^i_{[cls]}, \mathbf{s}^{\mathcal{V}}, \check{\mathbf{A}}^i \right] \in \mathbb{R}^{(k^i+2) \times d}. 
    \vspace{-0.1cm}
\end{equation}
These updated latent expressions are fed into the next layer.

\begin{table*}[t]
    \centering
    \footnotesize
    \scalebox{1.0}{
    \begin{tabular}{l|c|c|ccc|ccc|ccc}
    \hline
    \multirow{2}{*}{Methods}  & \multicolumn{2}{c|}{Encoders}& \multicolumn{3}{c|}{RefCOCO}& \multicolumn{3}{c|}{RefCOCO+} & \multicolumn{3}{c}{RefCOCOg} \\ \cline{2-12}
         & Visual & Textual & val & testA & testB & val & testA & testB & val(U) & test(U) & val(G) \\ \hline
    \multicolumn{12}{c}{\textit{Trained on each \textbf{Single RefCOCO Dataset}}} \\ 
    \hline
            \multicolumn{12}{l}{\textit{~~~Dual-encoder based Methods}} \\ \cdashline{1-12}[0.2pt/1pt]
    VPD\textsuperscript{23}~\cite{vpd} & VQGAN & CLIP & 75.67 & 77.39 & 73.23 & 67.98 & 71.82 & 60.39 & 66.42 & 66.75 & - \\
    CGFormer\textsuperscript{23}~\cite{late_9_cgformer} & Swin-B & BERT & 76.93 & 78.70 & 73.32 & 68.56 & 73.76 & 61.72 & 67.57 & 67.83 & 65.79 \\
    RISCLIP\textsuperscript{24}~\cite{joint_4_risclip} & CLIP-B & CLIP & 75.68 & 78.01 & 72.46 & 69.16 & 73.53 & 60.68 & 67.61 & 67.95 & - \\
    ReMamber\textsuperscript{24}~\cite{remamber} & VMamba-B & CLIP & 74.90 & 77.60 & 71.20 & 66.97 & 71.97 & 58.51 & 65.88 & 66.20 & - \\
    \hline
    \multicolumn{12}{l}{\textit{~~~Single-encoder based Methods}} \\ \cdashline{1-12}[0.2pt/1pt]
    Shared-RIS\textsuperscript{24}~\cite{sharedris}  & \multicolumn{2}{c|}{BEiT3-B} & 75.50 & 76.66 &73.03 & 70.34 & 73.75 & 65.07 & 68.50 & 69.17 & \underline{66.65} \\
    One-Ref\textsuperscript{24}~\cite{oneref}  & \multicolumn{2}{c|}{BEiT3-B}& \underline{77.57} &	\underline{79.05} &	\underline{75.11} &	\underline{71.25} &	\underline{75.41} &	\underline{65.45} &	\underline{69.37} &	\underline{69.70} & -  \\
    \hline
   Latent-VG (ours)  & \multicolumn{2}{c|}{BEiT3-B} & \textbf{77.72} & \textbf{79.52}  & \textbf{75.53} & \textbf{73.19}  & \textbf{76.24} & \textbf{67.35} & \textbf{71.11} & \textbf{71.49}  & \textbf{70.60} \\
   
    \hline
    \hline
    \multicolumn{12}{c}{\textit{Trained on \textbf{Combined RefCOCO Dataset}}} \\ 
    \hline
    
    \multicolumn{12}{l}{\textit{~~~Dual-encoder based Methods}} \\ 
    \cdashline{1-12}[0.2pt/1pt]
    PolyFormer\textsuperscript{23}~\cite{polyformer} & Swin-B & BERT & 75.96 & 77.09 & 73.22 & 70.65 & 74.51 & 64.64 & 69.36 & 69.88  & -\\
    RISCLIP\textsuperscript{24}~\cite{joint_4_risclip} & CLIP-B & CLIP & 76.01 & 78.63 & 71.94 & 69.67 & 74.30 & 61.37 & 69.61 & 69.56 & - \\
    ReMamber\textsuperscript{24}~\cite{remamber} & VMamba-B & CLIP & 77.77 & 79.81 & 75.57 & 69.82 & 73.93 & 63.08 & 70.96 & 71.81 & - \\
    EEVG\textsuperscript{24} & ViT-B & BERT & 79.49 & 80.87 & 77.39 & 71.86 & 76.67 & 66.31 & 73.56 & 73.47 & - \\
    \hline
    \multicolumn{12}{l}{\textit{~~~SAM based Methods}} \\ 
    \cdashline{1-12}[0.2pt/1pt]
    Chen. et al\textsuperscript{24}~\cite{sam_swin} & SAM \scriptsize + Swin-B & BERT & 77.14 & 78.33 & 74.33 & 71.75 & 75.70 & 65.69 & 70.72 & 71.43 & - \\
    Prompt-RIS\textsuperscript{24}~\cite{prompt_ris} & SAM \scriptsize + CLIP-B & CLIP &	78.10 & 81.21 & 74.64 & 71.13 & 76.60 & 64.25 & 69.17 & 70.47 & \underline{71.29} \\
    \hline
    \multicolumn{12}{l}{\textit{~~~LLM based Methods}} \\ 
    \cdashline{1-12}[0.2pt/1pt]
    GSVA-7B\textsuperscript{24}~\cite{gsva} & SAM \scriptsize + CLIP-L & Vicuna & 77.52 & 79.29 & 74.45 & 67.16 & 71.06 & 61.58 & 72.76 & 73.27 & -
 \\
    LaSagnA-7B\textsuperscript{24}~\cite{lasagna} & SAM \scriptsize + CLIP-L & Vicuna & 77.87 & 79.76 & 75.07 & 68.86 & 72.72 & 62.94 & 71.29 & 72.31 & -
 \\
    \hline
    \multicolumn{12}{l}{\textit{~~~Single-encoder based Methods}} \\ \cdashline{1-12}[0.2pt/1pt]
    One-Ref\textsuperscript{24}~\cite{oneref}  & \multicolumn{2}{c|}{BEiT3-B}& \underline{79.83} & \underline{81.86} & \underline{76.99} & \underline{74.68} & \underline{77.90} & \underline{69.58} & \underline{74.06} & \underline{74.92} & -  \\
    \hline
   Latent-VG (ours)  & \multicolumn{2}{c|}{BEiT3-B} & \textbf{81.01}  & \textbf{82.26} & \textbf{79.77} & \textbf{76.92} & \textbf{79.48} & \textbf{72.95} & \textbf{76.10} & \textbf{76.51} & \textbf{75.08} \\
    \hline
    \end{tabular}
    }
    \vspace{-0.2cm}
        \caption{mIoU comparison with various \textbf{RIS} methods on RefCOCO, RefCOCO+, and RefCOCOg datasets.
    U and G in RefCOCOg indicate UMD and Google partitions. 
    The best results are highlighed in \textbf{bold}, and the second-best are \underline{underlined}. 
    }
\label{tab:miou_ris}
\end{table*}

\begin{table*}[t]
    \centering
    \footnotesize

    \scalebox{1.0}{
    \begin{tabular}{l|c|ccc|ccc|ccc}
    \hline
    \multirow{2}{*}{Methods}  & \multirow{2}{*}{Backbone}& \multicolumn{3}{c|}{val}& \multicolumn{3}{c|}{testA} & \multicolumn{3}{c}{testB} \\ 
    \cline{3-11}
         & & mIoU & oIoU & N-acc. & mIoU & oIoU & N-acc. & mIoU & oIoU & N-acc. \\ \hline
    ReLA\textsuperscript{23}~\cite{late_1_ReLA} & Swin-B  &	63.60 & 62.42 & 56.37 & 70.03 & 69.26 & 59.02 & 61.02 & 59.88 & 58.40 \\
    EEVG\textsuperscript{24}~\cite{eevg} & ViT-B & 62.75 & 64.04 & - & 70.93 & \underline{71.65} & - & 62.79 & 62.77 & - \\
    MABP\textsuperscript{24}~\cite{mabp} & Swin-B  & \underline{68.86}  & \underline{65.72} & 62.18 & \underline{72.81} & 71.59 & - & \underline{64.04} & 62.76 & - \\
    HDC\textsuperscript{24}~\cite{hdc}  & Swin-B  &	68.23 & 65.42 & \underline{63.38} & 72.52 & 71.60 & 65.29 & 63.85 & \underline{62.79} & \underline{60.68} \\
    \hline
    LISA\textsuperscript{24}~\cite{lisa} & SAM \scriptsize + CLIP-L & 61.76 & 61.63 & 54.67 & 66.27 & 68.50 & 50.01 & 58.84 & 60.63 & 51.91
 \\
    GSVA-7B\textsuperscript{24}~\cite{gsva} & SAM \scriptsize + CLIP-L& 66.47 & 63.29 & 62.43 & 71.08 & 69.93 & \underline{65.31} & 62.23 & 60.47 & 60.56
 \\
    \hline
   Latent-VG (ours)  & BEiT3-B & \textbf{72.45} & \textbf{68.23} & \textbf{70.42} & \textbf{74.51} & \textbf{73.53} & \textbf{68.64} & \textbf{66.12} & \textbf{64.16} & \textbf{62.34} \\
    \hline
    \end{tabular}
    }
    \vspace{-0.2cm}
        \caption{Comparison with various \textbf{GRES} methods on gRefCOCO dataset. 
    }
    \vspace{-0.4cm}
\label{tab:gres}
\end{table*}

\subsection{Learning Objective}
\label{subsec:learning objective}
Our loss function consists of two components: 1) a proposed positive-margin contrastive loss $\mathcal{L}_{\text{pos-cont}}$ to align multiple generated latent expressions with the original text while preserving diversity among the variants, and 2) segmentation loss $\mathcal{L}_{\text{seg}}$ for pixel-level binary classification.
The final loss function is defined as $\mathcal{L} = \mathcal{L}_{\text{pos-cont}} + \mathcal{L}_{\text{seg}}$.

\vspace{-0.2cm}
\paragraph{Positive-Margin Contrastive Loss.}
The na\"ive contrastive learning (\ie, InfoNCE~\cite{cpc} or Triple~\cite{triplet}) forces positive samples into strict alignment, which could lead to nearly identical representations with the original text or the loss of novel semantic deviations in latent textual representations.
To address these issues, we propose a positive-margin contrastive loss, allowing a margin of fluctuations between positive samples for diverse representations:
\vspace{-0.1cm}
\begin{align}
    &\mathcal{L}_{\text{pos-cont}} = -\frac{1}{N}\sum_{i=1}^{N}\text{log} \frac{\text{exp}(\text{min}(1, \gamma + s_{i})/\tau)}{\sum_{k\in{\mathcal{N}_{i}}}\text{exp}(s_{k}/\tau)}, \\
     &\text{~~~~~~~~~where~~}  s_i = \mathbf{t}_{o}^{\top} {\mathbf{z}}_{o}^i /(\| \mathbf{t}_{o}\|\|{\mathbf{z}}_{o}^i\|), \nonumber
    \vspace{-0.1cm}
\end{align}
$\mathbf{t}_o=\text{MLP}(\mathbf{t}_{[cls],L})$, $\mathbf{z}_o^i=\text{MLP}(\mathbf{z}^i_{[cls],L})$,  $\gamma$ is a margin, $\mathcal{N}_{i}$ is the set of negative samples for $i$-th positive (\ie, class tokens of latent expressions from other batches), and $\tau$ is a temperature.

\vspace{-0.2cm}
\paragraph{Segmentation Loss.}
The final probability map $\hat{\mathbf{p}}$ is generated by averaging all similarity maps computed between the upsampled visual feature map $\mathbf{F}^{\mathcal{V}}$ and each embedded class token in all expressions,  $\mathbf{t}_{[cls]}$ and  $\mathbf{z}_{[cls]}^i$.
\vspace{-0.1cm}
\begin{equation}
    \hat{\mathbf{p}} = \frac{\sigma(\mathbf{F}^{\mathcal{V}} \cdot \mathbf{t}_{o}) + \sum_{i=1}^{N} \sigma ( \mathbf{F}^{\mathcal{V}} \cdot {\mathbf{z}}_{o}^{i})}{N+1},
    \vspace{-0.1cm}
\end{equation}
where $\mathbf{F}^{\mathcal{V}} \in \mathcal{R}^{h/4 \times w/4}$ is the twice upsampled and reshaped visual patches via a 2-layer deconvolution, $\sigma(\cdot)$ is the sigmoid function.
We apply a segmentation loss, comprising a binary cross-entropy loss and a dice loss~\cite{dice_loss}), to the predicted map $\hat{\mathbf{p}}$ using a ground truth mask $\mathbf{p}$, as follows: $\mathcal{L}_{\text{grd}} = \lambda_\text{bce}\mathcal{L}_{\text{bce}}(\hat{\mathbf{p}}, \mathbf{p}) + \lambda_\text{dice}\mathcal{L}_{\text{dice}}(\hat{\mathbf{p}}, \mathbf{p})$.

At the inference step for RIS, the segmentation mask is generated by thresholding $\hat{\mathbf{p}}$ at a predefined value and resizing it to the target image dimensions $(h, w)$. 
For REC, unlike previous methods that employ a separate detection decoder, we derive the bounding box by drawing a tight-fitting box around the resulting RIS mask.

\section{Experiments}
\label{sec:experiments}

\subsection{Implementation Details}
Our framework is built on the pre-trained BEiT-3~\cite{beit3}  with a ViT-B model size, which has a patch size $p=16$, $L=12$ layers, and a channel dimension of $d=768$.
For latent expressions, we use $N=2$, $p_i=\{0.2, 0.15\}$, and $k^i=\{4, 6\}$.
In the visual concept injector, the threshold $\kappa$ is set to the mean similarity value, and $N_c=100$ is used.
As in previous works~\cite{late_3_cris, late_1_ReLA}, the input image size is set to $(480, 480)$, and a feature pyramid network~\cite{vitadapter} is employed.
Our framework is trained with an AdamW optimizer with a batch size of 64 on four A6000 GPUs.
For a loss function, $\lambda_{\text{bce}}=2$, $\lambda_{\text{dice}}=0.5$, and $\gamma=0.2$ are used.
At inference, all dropout probabilities $\{p_i\}_{i=1}^{N}$ are set to zero and a threshold of 0.35 is applied for a mask generation. 


\subsection{Dataset and Metric}
\paragraph{Dataset.}
We evaluate our model on various benchmark datasets for visual grounding tasks (RIS, REC, and GRES): RefCOCO(+/g) datasets~\cite{refcoco, refcoco_umd, refcocog_google} for RIS, REC, and a gRefCOCO dataset~\cite{late_1_ReLA} for GRES. 
The detailed dataset information is provided in the supplementary materials.


\vspace{-0.3cm}
\paragraph{Metric.}
For RIS, we use the standard metrics: mIoU, oIoU, and prec@X; mIoU (mean IoU) is the average IoU across all samples, while oIoU (overall IoU) divides the total intersection by the total union over all samples. Prec@X is the percentage of samples with IoU above threshold X.
For REC, we employ Prec@50 (Acc.), which is an accuracy only when IoU is over 0.5.
For GRES, in addition to mIoU and oIoU, we also compute N-acc (No-target accuracy) to evaluate the accuracy on no-target samples.
N-acc is defined as $\frac{TP}{TP+FN}$, where a true positive (TP) indicates correctly predicting no target, otherwise a false negative (FN).

\subsection{Main Result}
\paragraph{Referring Image Segmentation.}
In Tab.~\ref{tab:miou_ris}, we compare the mIoU results of our Latent-VG with state-of-the-art (SoTA) RIS methods on the RIS benchmark under two settings: (1) in the \textbf{single dataset setting}, where each model is trained on an individual RefCOCO dataset, our method achieves SoTA performance across all dataset splits;
 and (2) in the \textbf{combined dataset setting}, where the methods are trained on the combined RefCOCO(+/g) datasets, our Latent-VG surpasses the SoTA methods, including One-Ref~\cite{oneref} which shares the same backbone, BEiT3-B, as ours.
Notably, our approach shows strong performance on RefCOCO+ and RefCOCOg, which offer more complicated textual descriptions than RefCOCO, {implying that our latent expressions are effective not only for brief input text but also for challenging textual forms.}
The oIoU and efficiency comparisons are reported in the supplementary materials.



\begin{table*}[t]
    \centering
    \footnotesize
    \scalebox{1.0}{
    \begin{tabular}{l|c|c|c|ccc|ccc|cc}
    \hline
    \multirow{2}{*}{Methods}  & \multicolumn{2}{c|}{Encoders}&  \multirow{2}{*}{\makecell{Data\\Size}}  & \multicolumn{3}{c|}{RefCOCO}& \multicolumn{3}{c|}{RefCOCO+} & \multicolumn{2}{c}{RefCOCOg} \\ \cline{2-3} \cline{5-12}
         & Visual & Textual & & val & testA & testB & val & testA & testB & val & test \\ \hline
    \multicolumn{12}{c}{\textit{Trained on each \textbf{Single RefCOCO Dataset}}} \\ 
    \hline
            \multicolumn{12}{l}{\textit{~~~Dual-encoder based Methods}} \\ \cdashline{1-12}[0.2pt/1pt]
    TransVG++\textsuperscript{23}~\cite{transvg++} & ViT-B & BERT & S-Ref & 86.28 & 88.37 & 80.97 & 75.39 & 80.45 & 66.28 & 76.18 & 76.30
  \\
  Dyn.MDETR\textsuperscript{23}~\cite{dynamicmdetr} & CLIP-B & CLIP & S-Ref & 85.97 & 88.82 & 80.12 & 74.83 & 81.70 & 63.44 & 74.14 & 74.49 \\
    HiVG\textsuperscript{24}~\cite{hivg} & CLIP-B & CLIP & S-Ref & 87.32 & 89.86 & 83.27 & 78.06 & 83.81 & 68.11 & 78.29 & 78.79  \\
    EEVG\textsuperscript{24}~\cite{eevg} & ViT-B & BERT & S-Ref &88.08 & 90.33 & \underline{85.50} & 77.97 & 82.44 & 69.15 & 79.60 & 80.24 \\
    \hline
    \multicolumn{12}{l}{\textit{~~~Single-encoder based Methods}} \\ \cdashline{1-12}[0.2pt/1pt]
    One-Ref\textsuperscript{24}~\cite{oneref}  & \multicolumn{2}{c|}{BEiT3-B}& S-Ref & \underline{88.75} & \underline{90.95} & 85.34 & \underline{80.43} & \underline{86.46} & \underline{74.26} & \textbf{83.68} & \textbf{83.52}  \\
    \hline
   Latent-VG (ours)  & \multicolumn{2}{c|}{BEiT3-B} & S-Ref & \textbf{89.75} & \textbf{92.42} & \textbf{85.91} & \textbf{83.18} & \textbf{87.71} & \textbf{76.48} & \underline{82.56} & \underline{83.13} \\

    \hline
    \hline
    \multicolumn{12}{c}{\textit{Trained on \textbf{Combined RefCOCO Dataset}}} \\ 
    \hline
    
    \multicolumn{12}{l}{\textit{~~~Dual-encoder based Methods}} \\ 
    \cdashline{1-12}[0.2pt/1pt]
    PolyFormer\textsuperscript{23}~\cite{polyformer} & Swin-B & BERT & C-Ref & 89.73 & 91.73 & 86.03 & 83.73 & 88.60 & 76.38 & 84.46 & 84.96 \\
    EEVG\textsuperscript{24}~\cite{eevg} & ViT-B & BERT & C-Ref & 90.47 & 92.73 & 87.72 & 81.79 & 87.80 & 74.94 & 85.19 & 84.72 \\
    HiVG\textsuperscript{24}~\cite{hivg} & CLIP-B & CLIP & C-Ref,ReferIt,Flickr & 90.56 & 92.55 & 87.23 & 83.08 & 87.83 & 76.68 & 84.71 & 84.69 \\
    Grounding-DINO\textsuperscript{24}~\cite{grounding_dino} & Swin-L & BERT & C-Ref,GoldG,\textit{etc} & 90.56 & 93.19 & 88.24 & 82.75 & 88.95 & 75.92 & 86.13 & 87.02 \\
    \hline
    \multicolumn{12}{l}{\textit{~~~LLM based Methods}} \\ 
    \cdashline{1-12}[0.2pt/1pt]
    Ferret-7B\textsuperscript{24}~\cite{ferret} & CLIP-L & Vicuna &  GRIT~\cite{ferret} & 87.49 & 91.35 & 82.45 & 80.78 & 87.38 & 73.14 & 83.93 & 84.76
 \\
    LION-12B\textsuperscript{24}~\cite{lion} & EVA-G & FlanT5 & VG,COCO,\textit{etc} &89.80 & 93.02 & 85.57 & 83.95 & 89.22 & 78.06 & 85.52 & 85.74
 \\
    \hline
    \multicolumn{12}{l}{\textit{~~~Single-encoder based Methods}} \\ \cdashline{1-12}[0.2pt/1pt]
    Sim-VG\textsuperscript{24}~\cite{simvg}  & \multicolumn{2}{c|}{BEiT3-B} & C-Ref,ReferIt,Flickr & 90.59 & 92.80 & 87.04 & 83.54 & 88.05 & 77.50 & 85.38 & 86.28 \\
    One-Ref\textsuperscript{24}~\cite{oneref}  & \multicolumn{2}{c|}{BEiT3-B} & C-Ref,ReferIt & \textbf{91.89} & \underline{94.31} & \underline{88.58} & \underline{86.38} & \underline{90.38} & \underline{79.47} & \underline{86.82} & \textbf{87.32} \\
    \hline
   Latent-VG (ours)  & \multicolumn{2}{c|}{BEiT3-B} & C-Ref & \underline{91.75} & \textbf{94.64} & \textbf{88.62} & \textbf{86.41} &  \textbf{90.57} & \textbf{80.59} & \textbf{87.01} & \underline{87.11} \\
    \hline
    \end{tabular}
    }
    \vspace{-0.2cm}
        \caption{Comparison with various \textbf{REC} methods on RefCOCO, RefCOCO+, and RefCOCOg datasets. S-Ref denotes training on a single dataset, and C-Ref indicates the combined RefCOCO(+/g) dataset.
    }
    \vspace{-0.2cm}
\label{tab:rec}
\end{table*}

\vspace{-0.3cm}
\paragraph{Generalized Referring Expression Segmentation.}
In Tab.~\ref{tab:gres}, we present the performance of our method on the GRefCOCO dataset, which serves as the GRES benchmark.
To handle no-target cases in the GRES task, we only add an empty token for binary classification between empty and non-empty cases.
With this remarkably simple modification to our framework, our method outperforms all compared SoTA methods, indicating its superior generalization and robustness.
Further details on our modification for the GRES task are provided in our supplementary material.

\begin{table}[t]
    \centering
    \footnotesize
    \scalebox{1.00}{
    \begin{tabular}{l|c|c|c@{\hspace{5pt}}c@{\hspace{5pt}}}
    \hline
    {Methods}     &  SD & VCI & {mIoU} & {oIoU} \\ 
     \hline
     No Latent Exps &  &  & 68.76 \scriptsize\textcolor{gray}{+0.00} & 66.30 \scriptsize\textcolor{gray}{+0.00} \\ \hline
    \multirow{4}{*}{\makecell[l]{$+$ Latent Exps}} &   &  & 71.03 \scriptsize\textcolor{red}{+2.27} & 68.29 \scriptsize\textcolor{red}{+1.99}\\
    & \ding{51} &   & 71.59 \scriptsize\textcolor{red}{+2.83} & 68.81 \scriptsize\textcolor{red}{+2.51} \\
    &  & \ding{51} & 71.86 \scriptsize\textcolor{red}{+3.10} & 69.62 \scriptsize\textcolor{red}{+3.32} \\
    & \ding{51} & \ding{51} & 72.63 \scriptsize\textcolor{red}{+3.87} & 70.13 \scriptsize\textcolor{red}{+3.83} \\ \hline
    $+$ $\mathcal{L}_{\text{pos-cont}}$ & \ding{51} & \ding{51} & 73.19 \scriptsize\textcolor{red}{+4.43} & 70.92 \scriptsize\textcolor{red}{+4.62} \\
    \hline
    \end{tabular}
    }
    \vspace{-0.2cm}
        \caption{Ablation on proposed components. \textit{No Latent Exps} denotes the base model without any proposed methods. \textit{SD} and \textit{VCI} denote the subject distributor and the visual concept injector, respectively. 
        The analysis of parameters and FLOPs in each component are reported in our supplementary.
        }
        \vspace{-0.3cm}
    \label{tab:ablation}
\end{table}





\vspace{-0.3cm}
\paragraph{Referring Expression Comprehension.}
We also present the performance of our methods on the REC task under two settings analogous to those in the RIS task, as shown in Tab.~\ref{tab:rec}.
In both settings, our approach demonstrates superior performance.
In particular, even though Sim-VG~\cite{simvg} and One-Ref~\cite{oneref} are trained with additional REC datasets (\eg, ReferIt~\cite{referit} or Flickr30k~\cite{flickr30k}) in the combined setting, our Latent-VG outperforms them on most datasets {despite using less training data and without requiring any task-specific decoder.}
Moreover, our model surpasses LLM-based methods, such as Ferret-7B~\cite{ferret}, LION-12B~\cite{lion}, and Grounding-DINO~\cite{grounding_dino}, all of which utilize more extensive grounding data and have significantly larger model sizes.

\vspace{-0.3cm}

\subsection{Ablation Study}
\vspace{-0.2cm}
We conduct ablation studies on a validation split of RefCOCO+ to validate the effectiveness of our approach.

\vspace{-0.3cm}
\paragraph{Effects of Each Proposed Component.}
In Tab.~\ref{tab:ablation}, we analyze the contribution of each component in our framework.
Starting with the base model, not using latent expression generation (noted as the ``\textit{No Latent Exps}" row), we observe that introducing na\"ive latent expressions yields significant performance improvements.
Adding the subject distribution further improves performance by ensuring a consistent subject reference.
Guiding distinct visual details through attribute tokens provides additional gains.
Furthermore, the combination of these modules leads to better results, supporting our design of shared-subject and distinct-attributes.
Finally, the introduction of the positive-margin contrastive loss further enhances performance by {allowing for slight deviations to diversify latent representations rather than strict alignments.}
Detailed analyses of parameters and FLOPs increases for each component are provided in our supplementary; our proposed methods marginally increase computational costs (about 12M params and 3 GFLOPs).



\begin{table}[t]
    \centering
    \footnotesize
    \scalebox{1.00}{
    \begin{tabular}{c|l|ccc}
    \hline
    \# Latent Exps $N$ & \# Tokens $k^i$ & mIoU & oIoU & GFLOPs \\ \hline
    No Latent Exps & - & 68.76 & 66.30 & 195 \\
    \hline
    \multirow{2}{*}{1}     & $\{4\}$  & 71.72 & 69.26 & 197 \\ 
         & $\{10\}$  & 72.16 & 69.84 & 198 \\ \hline
         
    \multirow{3}{*}{\textbf{2}}     & $\{4, 4\}$ & 73.09 & 70.61  & 198 \\
     & $\textbf{\{4, 10\}}$ & \textbf{73.19} & \textbf{70.92} & 198 \\
         & $\{6, 14\}$ & 72.61 & 70.22  & 199  \\ \hline
    \multirow{2}{*}{3} & $\{4, 4, 4\}$ & 72.15 & 69.84 & 199 \\
    & $\{4, 10, 16\}$ & 72.45 & 69.97 & 200 \\
    \hline
    \end{tabular}
    }
    \vspace{-0.2cm}
        \caption{Ablation study on the number of latent expressions and the token length within each latent expression.}
        \vspace{-0.2cm}
    \label{tab:num_exp}
\end{table}





\begin{table}[t]
    \centering
    \footnotesize
    \begin{tabular}{@{\hspace{6pt}}l|c@{\hspace{6pt}}c@{\hspace{6pt}}c@{\hspace{6pt}}c@{\hspace{6pt}}c@{\hspace{6pt}}}
    \hline
     Methods     & mIoU & oIoU & Pr@0.5 & Pr@0.7 & Pr@0.9  \\ \hline 
     InfoNCE~\cite{cpc} & 72.03 & 69.72 & 83.25 & 74.79 & 24.60 \\ 
     Triplet~\cite{triplet} & 72.57 & 70.42 & \textbf{84.12} & 75.75 & 23.55 \\  
     ArcFace~\cite{arcface} & 72.73 & 70.43 & 83.93 &  76.24 & 24.54 \\
     \cdashline{1-6}[0.2pt/1pt]
    Positive-margin (ours)    & \textbf{73.19} & \textbf{70.68} & 83.88 & \textbf{76.32} & \textbf{26.45}  \\ 
    \hline
    \end{tabular}
    \vspace{-0.2cm}
    \caption{Comparisons with other contrastive learning objectives.}
    \label{tab:different_contrastive}
    \vspace{-0.4cm}
\end{table}
\begin{table}[]
    \centering
    \footnotesize
    \begin{tabular}{l|c@{\hspace{6pt}}c@{\hspace{6pt}}c@{\hspace{6pt}}c@{\hspace{6pt}}c@{\hspace{6pt}}}
    \hline
    Ablations     & mIoU & oIoU & Pr@0.5 & Pr@0.7 & Pr@0.9  \\ \hline
    \multicolumn{6}{l}{~~ (a) Dropout $p_i$ in \textit{Latent Exp Init}} \\ \cdashline{1-6}[0.2pt/1pt]
    No dropout & 72.33 & 70.30 & 83.49 & 75.31 & 23.69 \\
    $\{0.4, 0.3\}$ & 73.03 & 70.46 & 84.63 & 76.25& 24.14 \\  \cdashline{1-6}[0.2pt/1pt]
    $\textbf{\{0.2, 0.15\}}$ & \textbf{73.19} & \textbf{70.68} & \textbf{83.88} & \textbf{76.32} & \textbf{26.45}\\ 
    \hline
    \multicolumn{6}{l}{~~(b) Threshold $\kappa$ for target-related patches in \textit{Visual Concept Injector}} \\ \cdashline{1-6}[0.2pt/1pt]
     No threshold   & 72.03 & 69.42 & 83.26 & 75.08 & 24.15 \\ 
     Top-10 patches & 71.47 & 68.90 & 82.87 & 74.20 & 24.74 \\ 
     Top-30 patches & 72.47 & 69.88 & 83.49 & 75.81 & 24.60 \\  \cdashline{1-6}[0.2pt/1pt]
    \textbf{Mean value}& \textbf{73.19} & \textbf{70.68} & \textbf{83.88} & \textbf{76.32} & \textbf{26.45} \\
     \hline
     \multicolumn{6}{l}{~~ (c) \# Concept tokens $N_c$ in \textit{Visual Concept Injector}} \\ \cdashline{1-6}[0.2pt/1pt]
     50 & 72.95 & \textbf{70.69} & \textbf{84.46} & 75.60 & 24.84\\
      150 & 72.27 & 70.00 & 83.98 & 74.93 & 23.85 \\
     100 \scriptsize w/o ortho init~\cite{orthogonal_init} & 72.59 & 70.31 & 83.85 & 75.61 & 24.69 \\  \cdashline{1-6}[0.2pt/1pt]
     \textbf{100} & \textbf{73.19} & 70.68 & 83.88 & \textbf{76.32} & \textbf{26.45} \\
     \hline
     \multicolumn{6}{l}{~~ (d) Margin $\gamma$ in $\mathcal{L}_{\text{pos-cont}}$} \\ \cdashline{1-6}[0.2pt/1pt]
     No margin~\cite{cpc} & 72.03 & 69.72 & 83.25 & 74.79 & 24.60 \\
     0.1 & 72.94 & \textbf{71.11} & \textbf{84.46} & 75.63 & 23.55 \\
     0.3 & 72.85 & 70.85 & 84.13 & 76.23 & 24.29 \\ \cdashline{1-6}[0.2pt/1pt]
     \textbf{0.2} &\textbf{ 73.19 }& 70.68 & 83.88 & \textbf{76.32} & \textbf{26.45} \\

    \hline
    \end{tabular}
    \vspace{-0.2cm}
    \caption{Ablation study within proposed components. \textit{ortho init} indicates the orthogonal initialization~\cite{orthogonal_init}.}
    \vspace{-0.4cm}
    \label{tab:ablation_within_component}
\end{table}

\vspace{-0.3cm}
\paragraph{Number and Length in Latent Expressions.}
We examine the effect of the number ($N$) and length ($k^i$) of our generated latent expressions, as reported in Tab.~\ref{tab:num_exp}.
With $N=1$, our method improves performance over the baseline without latent expressions (``\textit{No Latent Exps}"), and using $N=2$ and $k^i=\{4,10\}$ yields the best results.
Especially, our experiments indicate that latent expressions with varied lengths (\eg, $\{4,10\}$) are more effective than those with equal lengths (\ie, $\{4,4\}$), highlighting that representing the target with different token lengths captures different details.
Moreover, too many or excessively long latent expressions decrease the performance compared to the best result, probably because too many latent tokens act as noise.

\vspace{-0.3cm}
\paragraph{Other Contrastive Learning Objectives.}
To validate the effectiveness of the proposed positive-margin contrastive learning in the generation of latent textual representations, we experiment with other contrastive objectives (\eg, InfoNCE~\cite{cpc}, Triplet~\cite{triplet}, and ArcFace~\cite{arcface} losses) in Tab.~\ref{tab:different_contrastive}.
Unlike the compared methods focusing on negative separation, our method allows variants in positive samples to obtain diversified target textual representations.
This leads our proposed loss function to the best performance.





\vspace{-0.3cm}
\paragraph{Ablation Study within Proposed Component.}
Tab.~\ref{tab:ablation_within_component} illustrates the effects of each module design and hyperparameter settings within the proposed components on performance.
Tab.~\ref{tab:ablation_within_component}a shows that applying dropout to an input text in the latent expression initialization improves performance, implying that reducing reliance on the original text is beneficial.
{In Tab.~\ref{tab:ablation_within_component}b, using the mean similarity threshold for selecting target-related patches in our \textit{visual concept injector} yields the best performance.
Using too few (\textit{e.g.}, Top-10 or Top-30) or too many (\textit{e.g.}, no threshold) patches results in marginal improvements, suggesting that the mean value effectively captures target-focused regions while avoiding irrelevant details or omissions.}
In Tab.~\ref{tab:ablation_within_component}c, using $N_c=100$ concept tokens achieves the best performance, and orthogonal initialization~\cite{orthogonal_init} outperforms random initialization.
Finally, as reported in Tab.~\ref{tab:ablation_within_component}d, setting the contrastive margin to $\gamma=0.2$ achieves better results than strict alignment without margin.









\begin{figure}[t]
    \centering
    \includegraphics[width=1\linewidth]{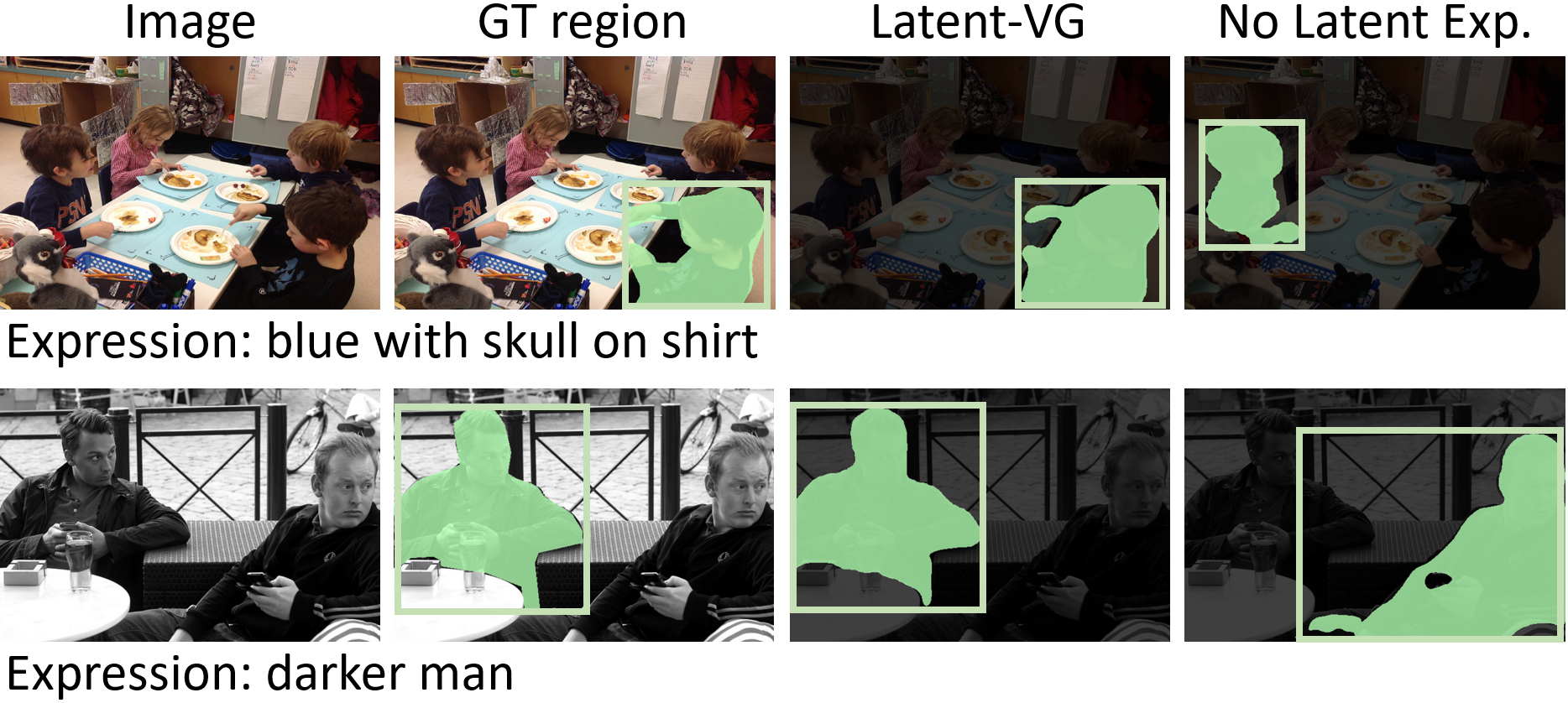}
    \vspace{-0.7cm}
    \caption{Qualitative analysis of Latent-VG compared to the baseline without any proposed methods (denoted as \textit{No Latent Exp.}).} 
    \label{fig:qual_no_exp} 
        \vspace{-0.1cm}
\end{figure}

\begin{figure}[t]
    \centering
    \includegraphics[width=1\linewidth]{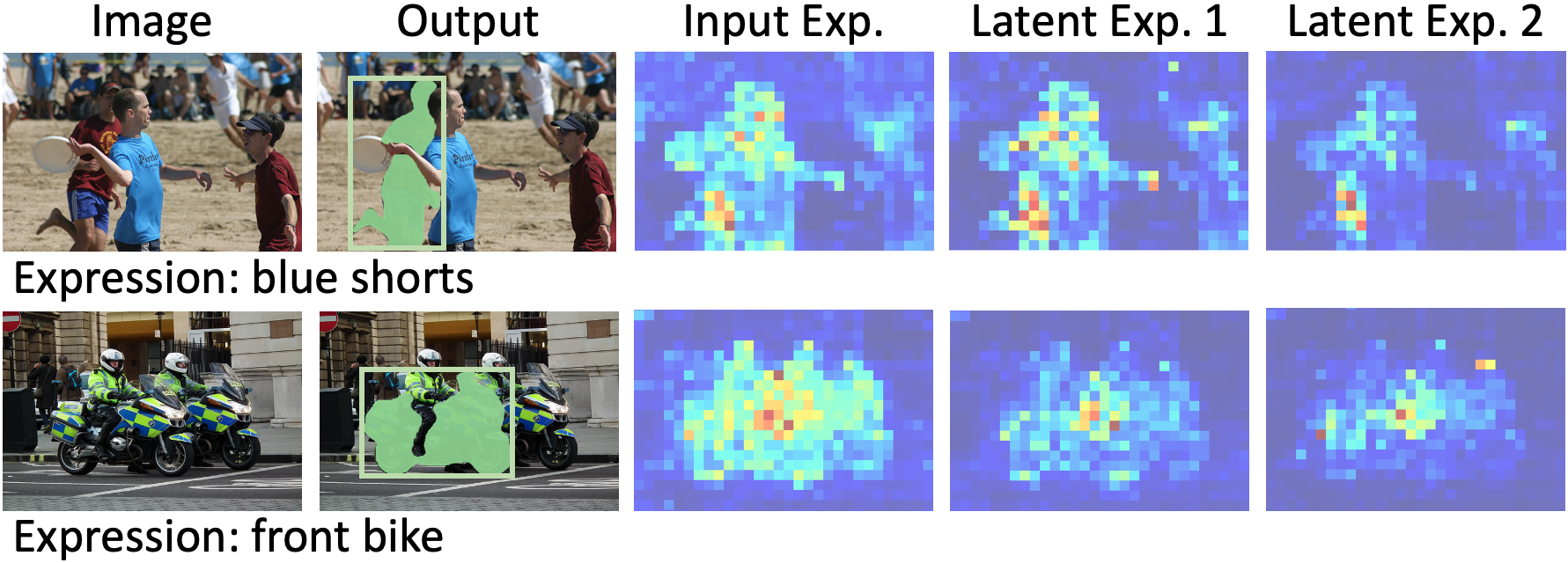}
    \vspace{-0.7cm}
    \caption{Visualization of attention maps on each expression.} 
    \label{fig:attn_maps} 
    \vspace{-0.3cm}
\end{figure}

\vspace{-0.05cm}
\subsection{Qualitative Analysis.}
Fig.~\ref{fig:qual_no_exp} visualizes the outputs of our Latent-VG compared to the base model without any of the proposed methods (termed as ``\textit{No Latent Exp.}").
By utilizing the latent expressions as additional target cues, our framework effectively handles ambiguous cases that contain similar objects to the target.
For instance, if ``\textit{blue with skull on shirt}" is given (in Fig.~\ref{fig:qual_no_exp}), the ``\textit{No Latent Exp}" model identifies the non-targeted left boy with the blue shirt.
While our approach correctly selects the target boy wearing a blue shirt.

\vspace{-0.2cm}
\paragraph{Attention Maps on each Expression.}
Fig.~\ref{fig:attn_maps} visualizes the attention maps on each expression.
In particular, each expression highlights different regions, indicating our approach goes beyond the semantics of the \textit{Input Exp.}, and references more collective cues via the \textit{Latent Exps}.
For example, in the first row, the \textit{Input Exp.} overly reveals the person regions, the \textit{Latent Exp. 1} puts a high weight on the hands of the middle person, and the \textit{Latent Exp. 2} focuses on the blue trousers.
In addition to the attention maps, further analysis on latent expressions, {including the qualitative results, the IoU scores, and the convergence of IoU scores for each expression, is discussed in the supplementary.}



\section{Conclusion}
\vspace{-0.1cm}
In this paper, we present a novel visual grounding framework that generates multiple latent expressions in the latent space from the original textual input.
By crafting the latent expressions to capture extra visual details of the target, we effectively address the limitations of visual grounding tasks (RIS and REC), which lack the comprehensive target details from the input expression provided.
Our framework not only achieves outstanding results on various visual grounding tasks, including RIS, REC, and GRES tasks, but also demonstrates that the generated latent expressions deliver additional cues to the model.

\paragraph{Acknowledgments.}
This work was supported by the NRF grants (RS-2021-NR05515 (10\%), RS-2024-00336576 (5\%), RS-2023-0022663 (5\%)), IITP grants (RS2022-II220926 (30\%), RS-2022-II220264 (5\%), RS-2024-00353131 (10\%)) funded by MSIT, and the KRIT grant funded by DAPA (Defense Acquisition Program Administration) (No. 21-107-E00-009-02) (35\%).
We also appreciate the high-performance GPU computing
support of HPC-AI Open Infrastructure via GIST SCENT.






{
    \small
    \bibliographystyle{ieeenat_fullname}
    \bibliography{main}
}
\clearpage
\appendix
\label{sec:appendix_section}
\setcounter{figure}{4}
\setcounter{table}{7}

\begin{strip}
\begin{center}
   {\Large \bf Latent Expression Generation for Referring Image Segmentation and Grounding \\
   - \textit{Supplementary materials} - \par}
   \vspace*{24pt}
   \large
   \lineskip .5em
   \vspace{-0.8cm}
\end{center}
\end{strip}

\begin{center}
    \large{\textbf{Table of Contents}}
\end{center}
We provide a table of contents for the supplementary:
\begin{enumerate}[label=\Alph*.]
    \item \hyperref[supple:dataset_details]{\textbf{Dataset Details}} 
    \begin{enumerate}[label=\Alph{enumi}.\arabic*.]
        \item RefCOCO
        \item RefCOCO+
        \item RefCOCOg
        \item GRefCOCO
    \end{enumerate}

    \item \hyperref[supple:gres_framework]{\textbf{Details on GRES Framework}} 
    \begin{enumerate}[label=\Alph{enumi}.\arabic*.]
        \item Framework
        \item Implementation Details
    \end{enumerate}

    \item \hyperref[supple:additional_experiments]{\textbf{Additional Experiments}} 
    \begin{enumerate}[label=\Alph{enumi}.\arabic*.]
        \item oIoU Results for RIS
        \item Efficiency Comparison
        \item Efficiency in each Proposed Component
    \end{enumerate}

    \item \hyperref[supple:additional_analysis]{\textbf{Analysis on Latent Expressions}} 
    \begin{enumerate}[label=\Alph{enumi}.\arabic*.]
        \item Qualitative Analysis of each Expression
        \item More Attention Maps on each Expression
        \item More Qualitative Analysis
        \item IoU Scores for each Expression
        \item Convergence of each IoU Score.
        \item Examples of an Extracted Subject
    \end{enumerate}

    \item \hyperref[supple:limitation]{\textbf{Further Discussion}}
    \begin{enumerate}[label=\Alph{enumi}.\arabic*.]
        \item Limitations
        \item Social Impact
    \end{enumerate}

    \item \hyperref[supple:more_visualization]{\textbf{Visualizations}}
    \begin{enumerate}[label=\Alph{enumi}.\arabic*.]
        \item Visualization on RIS and REC
        \item Visualization on GRES
    \end{enumerate}

\end{enumerate}

\section{Dataset Details}
\label{supple:dataset_details}
\subsection{RefCOCO}
RefCOCO~\cite{refcoco} is a dataset for referring image segmentation (RIS) and referring expression comprehension (REC), built on images and annotations (segmentation masks and bounding boxes) from MS-COCO~\cite{coco}. 
It was collected using the approach used in ReferItGame~\cite{referit}, where one player writes a referring expression for a segmented object, and another player selects the corresponding object in an image. 
The dataset contains 142,210 expressions for 50,000 objects across 19,994 images. 
It is divided into train, validation, and test sets. 
The test set is further split into testA and testB. 
The testA contains images with people, while the testB includes images with all other objects.
This split structure allows for separate evaluation of human and non-human referents.

\subsection{RefCOCO+}
RefCOCO+~\cite{refcoco} follows the same approach as RefCOCO but prohibits spatial terms like ``left" or ``right" and provides expressions based on object attributes.
Thus, RefCOCO+ is characterized as a more challenging dataset than RefCOCO, as it requires accurate object localization using only visual attributes without relying on positional cues. 
It includes 141,564 expressions for 49,856 objects across 19,992 images, and is split in the same way as RefCOCO.

\subsection{RefCOCOg}
RefCOCOg~\cite{refcocog_google, refcoco_umd}, unlike RefCOCO and RefCOCO+, was collected in a non-interactive setting via Amazon Mechanical Turk (AMT), resulting in the longer and more detailed textual forms. 
While RefCOCO and RefCOCO+ have concise expressions with an average length of 3.61 and 3.53 words, respectively, RefCOCOg expressions are significantly longer, averaging 8.43 words. 
This dataset was designed to evaluate a model’s ability to comprehend more complex and contextually rich referring expressions.
RefCOCOg consists of 95,010 expressions for 49,822 objects across 25,799 images and is divided into two partitions: the \textit{Google}~\cite{refcocog_google} split and the \textit{UMD}~\cite{refcoco_umd} split.
In the \textit{Google} split, objects are separately assigned to either the train or validation set while allowing the same image to appear in both sets without object overlaps.
The \textit{UMD} split separates the data into train, validation, and test sets.

\begin{figure}[b]
    \centering
      \includegraphics[width=0.95\linewidth]{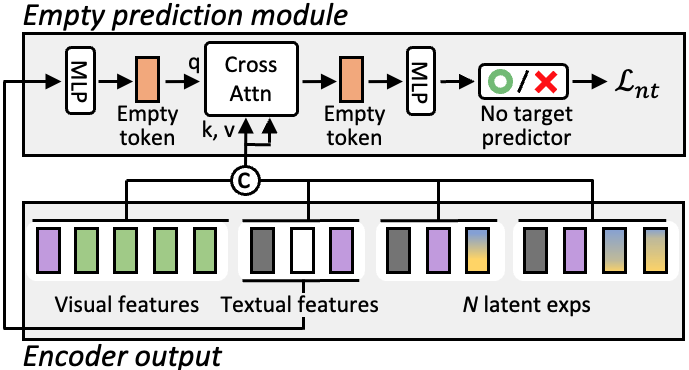}
    \caption{The illustration of our GRES framework, where the empty prediction module is added to the original framework after the feature extraction of the encoder. We also remove the subject distributor within an encoder. Other processes except for these minimal modifications are the same as the original one.} 
    \label{fig:gres_framework}
\end{figure}

\begin{table*}[t]
    \centering
    \footnotesize
    \scalebox{1.02}{
    \begin{tabular}{l|c|c|ccc|ccc|ccc}
    \hline
    \multirow{2}{*}{Methods}  & \multicolumn{2}{c|}{Encoders}& \multicolumn{3}{c|}{RefCOCO}& \multicolumn{3}{c|}{RefCOCO+} & \multicolumn{3}{c}{RefCOCOg} \\ \cline{2-12}
         & Visual & Textual & val & testA & testB & val & testA & testB & val(U) & test(U) & val(G) \\ \hline
    \multicolumn{12}{c}{\textit{Trained on each \textbf{Single RefCOCO Dataset}}} \\ 
    \hline
    \multicolumn{12}{l}{\textit{~~~Dual-encoder based Methods}} \\ 
    \cdashline{1-12}[0.2pt/1pt]
    VPD\textsuperscript{23}~\cite{vpd} & VQGAN & CLIP & 73.46 & 75.31 & 70.23 & 61.41 & 67.98 & 54.99 & 63.12 & 63.59 & - \\
    CGFormer\textsuperscript{23}~\cite{late_9_cgformer} & Swin-B & BERT & 74.75 & 77.30 & 70.64 & 64.54 & 71.00 & 57.14 & 64.68 & 64.09 & 62.51 \\
    RISCLIP\textsuperscript{24}~\cite{joint_4_risclip} & CLIP-B & CLIP & 73.57 & 76.46 & 69.76 & 65.53 & 70.61 & 55.49 & 64.10 & 65.09 & - \\
    ReMamber\textsuperscript{24}~\cite{remamber} & VMamba-B & CLIP & 74.54 & 76.74 & 70.89 & 65.00 & 70.78 & 57.53 & 63.90 & 64.00 & - \\
    \hline
    \multicolumn{12}{l}{\textit{~~~Single-encoder based Methods}} \\ \cdashline{1-12}[0.2pt/1pt]
    Shared-RIS\textsuperscript{24}~\cite{sharedris}  & \multicolumn{2}{c|}{BEiT3-B} & 75.50 & 76.66 &73.03 & 70.34 & 73.75 & \textbf{65.07} & 68.50 & 69.17 & \underline{66.65} \\
   One-Ref\textsuperscript{24}~\cite{oneref}  & \multicolumn{2}{c|}{BEiT3-B} & \textbf{77.55} & \textbf{80.96} & \underline{73.51} & \underline{70.82} & \underline{74.53}  & 64.06  & \underline{70.68} & \underline{70.61} &  - \\ 
   \hline
   Latent-VG (ours)  & \multicolumn{2}{c|}{BEiT3-B} & \underline{77.41} & \underline{79.92} & \textbf{74.83} & \textbf{70.92} & \textbf{74.56} & \underline{63.68}  & \textbf{70.74} & \textbf{70.82} & \textbf{69.19} \\

    \hline
    \hline
    \multicolumn{12}{c}{\textit{Trained on \textbf{Combined RefCOCO Dataset}}} \\ 
    \hline

    \multicolumn{12}{l}{\textit{~~~Dual-encoder based Methods}} \\ 
    \cdashline{1-12}[0.2pt/1pt]
    PolyFormer\textsuperscript{23}~\cite{polyformer} & Swin-B & BERT & 74.82 & 76.64 & 71.06 & 67.64 & 72.89 & 59.33 & 67.76 & 69.05 & -\\
    ReMamber\textsuperscript{24}~\cite{remamber} & VMamba-B & CLIP & 75.06 & 78.27 & 71.82 & 64.40 & 69.49 & 56.34 & 66.70 & 68.05 & - \\
    \hline
    \multicolumn{12}{l}{\textit{~~~SAM based Methods}} \\ 
    \cdashline{1-12}[0.2pt/1pt]
    Chen. et al\textsuperscript{24}~\cite{sam_swin} & SAM \scriptsize + Swin-B & BERT &  75.37 & 77.20 & 71.38 & 68.07 & 73.46 & 59.47 & 67.75 & 69.50 & - \\
    Prompt-RIS\textsuperscript{24}~\cite{prompt_ris} & SAM \scriptsize + CLIP-B & CLIP &76.36 & 80.37 & 72.29 & 67.06 & 73.58 & 58.96 & 64.79 & 67.16 & \underline{69.01} \\
    \hline
    \multicolumn{12}{l}{\textit{~~~LLM based Methods}} \\ 
    \cdashline{1-12}[0.2pt/1pt]
    GSVA-7B\textsuperscript{24}~\cite{gsva} & SAM \scriptsize + CLIP-L & Vicuna & 77.13 & 78.82 & 73.45 & 65.87 & 69.47 & 59.55 & 72.72 & 73.36 & - \\
    LaSagnA-7B\textsuperscript{24}~\cite{lasagna} & SAM \scriptsize + CLIP-L & Vicuna &76.30 & 77.38 & 72.76 & 64.42 & 67.62 & 58.63 & 71.13 & 72.01 & -  \\
    \hline
    \multicolumn{12}{l}{\textit{~~~Single-encoder based Methods}} \\ \cdashline{1-12}[0.2pt/1pt]
    One-Ref\textsuperscript{24}~\cite{oneref}  & \multicolumn{2}{c|}{BEiT3-B}& \textbf{81.06} & \textbf{83.05} & \underline{77.80} & \underline{72.24} & \underline{77.32} & \underline{67.08} & \underline{75.14} & \textbf{77.21}  & -  \\
    \hline
    Latent-VG (ours)  & \multicolumn{2}{c|}{BEiT3-B} & \underline{81.04} & \underline{82.67} & \textbf{79.77} & \textbf{75.27} & \textbf{78.25} & \textbf{69.65}  & \textbf{75.88} & \underline{76.55} & \textbf{75.02} \\
    \hline

    \hline
    \end{tabular}
    }
        \caption{oIoU comparison with other \textbf{RIS} methods on RefCOCO, RefCOCO+, and RefCOCOg datasets.
}
\label{tab:oiou_ris}
\end{table*}

\subsection{GRefCOCO}
GRefCOCO is a dataset designed for Generalized Referring Expression Segmentation (GRES)~\cite{late_1_ReLA}, extending the standard RefCOCO dataset by supporting the cases of multi-target and no-target expressions. 
Unlike traditional RIS datasets, where each expression corresponds to a single existent object, GRefCOCO offers a description referring to multiple or non-existent objects. 
This makes the dataset more flexible and suited for real-world scenarios.
The dataset consists of 278,232 referring expressions, including 80,022 multi-target expressions and 32,202 no-target expressions, covering 60,287 instances in 19,994 images.

\section{Details on GRES Framework}
\label{supple:gres_framework}
\subsection{Framework}

In Fig.~\ref{fig:gres_framework}, we present the GRES framework, where we add the empty prediction modules to the original model by (1) introducing an empty token to handle no-target cases, (2) interacting the empty token with the output of the encoder via cross-attention, and (3) imposing a binary classification loss ($\mathcal{L}_{nt}$) on the empty prediction.
We also remove a subject distributor within an encoder.
All other processes are the same as the original framework.

\subsection{Implementation Details for GRES}
The differences in the detailed implementations for a GRES framework lie in (1) halving the learning rate from 0.0001 to 0.00005 for more stable training on complex GRES scenarios, and (2) applying the loss weight of 0.5 to the empty binary classification objective ($\mathcal{L}_{nt}$).
Other training recipes are identical to the original visual grounding framework.



\section{Additional Experiments}
\label{supple:additional_experiments}

\subsection{oIoU Results for RIS.}
Tab.~\ref{tab:oiou_ris} shows the oIoU performance of our Latent-VG compared to SoTA RIS methods on the RIS benchmarks.
In the single dataset setting, we achieve superior performance over the methods~\cite{sharedris, oneref} based on the same backbone (\ie, BEiT3-B) as ours.
In the combined dataset setting, our Latent-VG surpass Prompt-RIS~\cite{prompt_ris} and LaSagnA-7B~\cite{lasagna}, even though they employ larger backbones (e.g., SAM and CLIP-L) than ours.


\subsection{Efficiency Comparison.}
In Tab.~\ref{tab:efficiency_total}, we compare the efficiency of our methods with other visual grounding methods.
The methods~\cite{early_3_dmmi, late_9_cgformer, joint_4_risclip, transvg++, mdetr, grounding_dino} utilizing multi-scale features in their decoder incur high FLOPs, whereas our simple decoding yields lower FLOPs over them with the best results.
Although One-Ref~\cite{oneref} for a REC model is more efficient than ours, it does not simultaneously perform a RIS task.


\begin{table}[h]
    \centering
    \footnotesize
    \begin{tabular}{l|cc|cc}
    \hline
    \multicolumn{5}{c}{~~~\textit{RIS Methods}} \\ \cdashline{1-5}[0.2pt/1pt]
    Methods & Params & GFLOPs & mIoU & oIoU \\ \hline
    
    DMMI~\cite{early_3_dmmi}     & 341M & 392 & 67.51 & 63.98  \\
    CGFormer~\cite{late_9_cgformer}     &  252M & 949 & 68.56 & 64.54 \\
    RISCLIP$^{\dagger}$~\cite{joint_4_risclip} & 375M & 1380 & 69.16 & 65.53 \\
    Shared-RIS$^{\dagger}$~\cite{sharedris} & 239M & 155 & 70.34 & 68.42 \\
    One-Ref$^{\dagger}$~\cite{oneref} & 267M & - & 71.25 & 70.82 \\
    \hline
    Latent-VG (ours) & 267M & 198 & \textbf{73.19} & \textbf{70.92}  \\
    \hline
    \hline

    \multicolumn{5}{c}{~~~\textit{REC Methods}} \\ \cdashline{1-5}[0.2pt/1pt]
    Methods & Params & GFLOPs &\multicolumn{2}{c}{Acc.} \\ \hline
        TransVG++$^{\dagger}$~\cite{transvg++}     & 206M & 396 &  \multicolumn{2}{c}{75.39}  \\
        MDETR$^{\ddagger}$~\cite{mdetr} & 185M & 642  & \multicolumn{2}{c}{81.13} \\
    Grounding-DINO$^{\ddagger}$~\cite{grounding_dino} & 342M & 464 & \multicolumn{2}{c}{82.75}  \\
    One-Ref$^{\dagger}$$^{\ddagger}$~\cite{oneref} & 234M & 162 & \multicolumn{2}{c}{86.38} \\
    \hline
    Latent-VG (ours) & 267M & 198 & \multicolumn{2}{c}{\textbf{86.41}}  \\
    \hline
    \end{tabular}
    \caption{Efficiency and performance comparison with other RIS and REC methods on the validation set of RefCOCO+. $\dagger$ denotes that the public code is not released. $\ddagger$ indicates the models trained on additional grounding data (\eg, Flickr30k or ReferIt) than ours.}
        \vspace{-0.2cm}
    \label{tab:efficiency_total}
\end{table}

\subsection{Efficiency of Proposed Components.}
In Tab.~\ref{tab:ablation_flops}, we analyze the incremental computational cost introduced by each proposed module.
Adapting the latent expression initializer increases the computational cost by 11M parameters and 2 GFLOPs, due to (1) length transform layers $\{\phi_i\}_{i=1}^{N}$ for initializing latent attributes and (2) additional MLP heads for processing each class token during prediction.
In contrast, adding the subject distributor incurs only a negligible increase in computational cost, with the additional parameters and GFLOPs being significantly lower than those of other modules.
Incorporating the visual concept injector adds 1M parameters and 1 GFLOPs since the parameters of the concept tokens and FLOPs for handling concept tokens are introduced.
The computation required for the loss function applied between class tokens is also negligible.

\begin{table}[h]
    \centering
    \footnotesize
    \scalebox{1.00}{
    \begin{tabular}{l|c|c|c@{\hspace{5pt}}c@{\hspace{5pt}}}
    \hline
    {Methods}     &  SD & VCI & {Params} & {GFLOPs} \\ \hline
     No Latent Exps &  &  & 255M & 195 \\ \hline
    \multirow{4}{*}{\makecell[l]{$+$ Latent Exps}} &  & & 266M & 197 \\
    & \ding{51} & & 266M  & 197  \\
    &  & \ding{51} & 267M  & 198 \\
    & \ding{51} & \ding{51} & 267M & 198 \\ \hline
    $+$ $\mathcal{L}_{\text{pos-cont}}$ & \ding{51} & \ding{51} & 267M & 198  \\
    \hline
    \end{tabular}
    }
    \caption{Analysis of the computational cost in different components. \textit{No Latent Exps} means the base model without any proposed methods. \textit{SD} and \textit{VCI} denote the subject distributor and the visual concept injector, respectively.}
    \label{tab:ablation_flops}
\end{table}






\section{Analysis on Latent Expressions}
\label{supple:additional_analysis}

\subsection{Qualitative Analysis of each Expression}
\begin{figure}[t]
    \centering
      \includegraphics[width=1\linewidth]{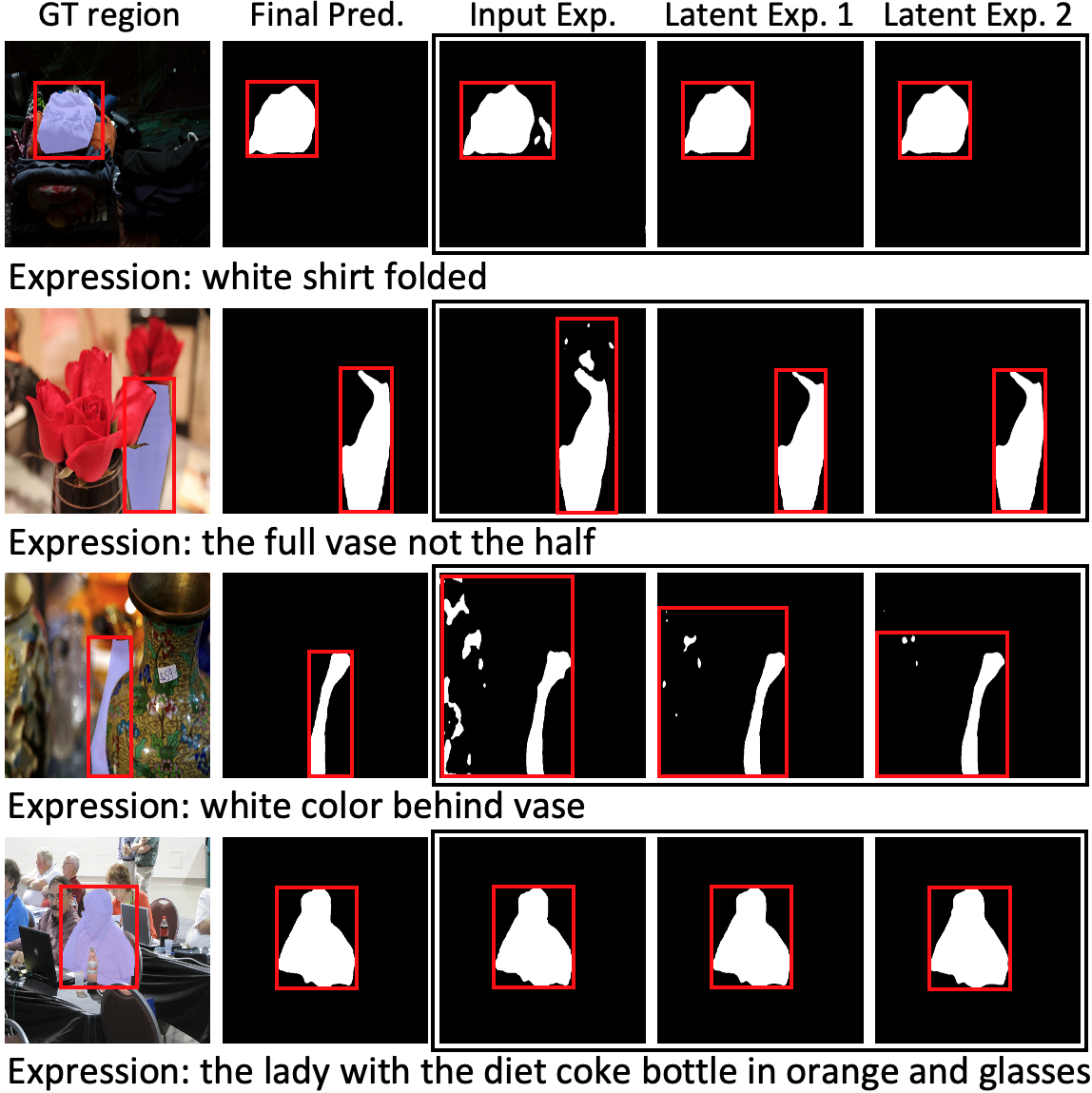}
    \caption{Qualitative analysis of each expression. The outputs for each expression (\ie, \textit{Input Exp.}, \textit{Latent Exp.1} and \textit{Latent Exp.2}) are obtained by thresholding the corresponding probability map before averaging them for the final prediction.} 
    \label{supfig:qual_each_exp}
\end{figure}

In Fig.~\ref{supfig:qual_each_exp}, we visualize the prediction result of each latent expression as well as the averaged final prediction with diverse cases.
Each prediction of individual expression (\ie, \textit{Input Exp.}, \textit{Latent Exp.1} and \textit{Latent Exp.2}) is obtained by thresholding the corresponding probability map before averaging them for the final prediction.
In the first and second rows, we present the cases where the noisy output of the \textit{Input Exp.} is complemented by the more precise output from the \textit{Latent Exps}, resulting in a correct final prediction.
In the third row, all predictions of each expression are noisy, yet the final averaged output is accurate.
In the last row, we visualize a result that all outputs are generated precisely.

\begin{figure}[t]
    \centering
      \includegraphics[width=1\linewidth]{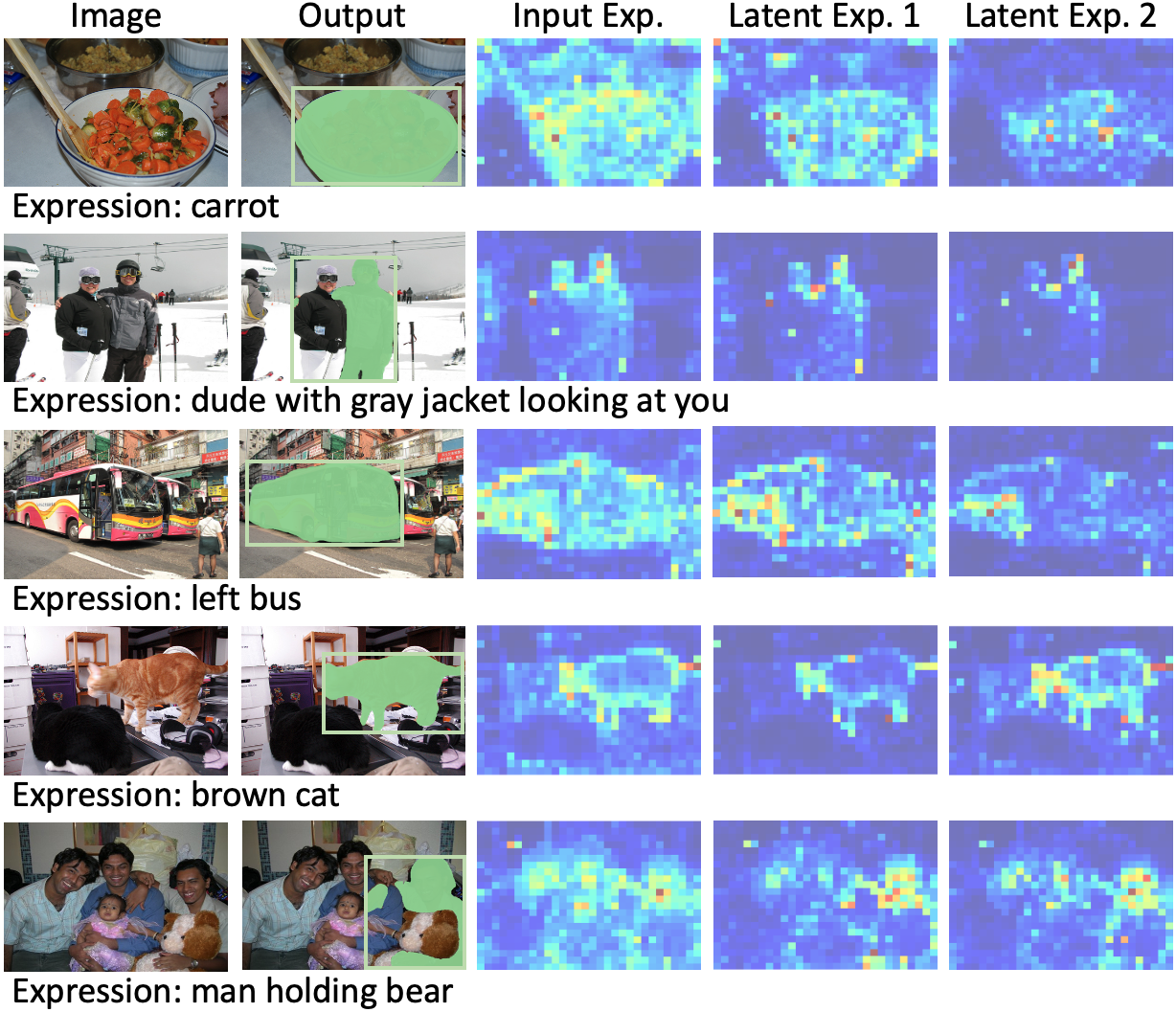}
    \caption{More attention maps on each expression.} 
    \label{supfig:more_attention}
\end{figure}




\subsection{More Attention Maps on each Expression}
Fig.~\ref{supfig:more_attention} presents additional attention maps for each expression.
To generate these maps, we average the attention scores from all self-attention layers within an encoder. 
The variability in the locations of the maximum attention weight, along with the differently activated regions, indicates that each expression exhibits distinct visual details.


\subsection{More Qualitative Analysis}
\begin{figure}[t]
    \centering
      \includegraphics[width=1\linewidth]{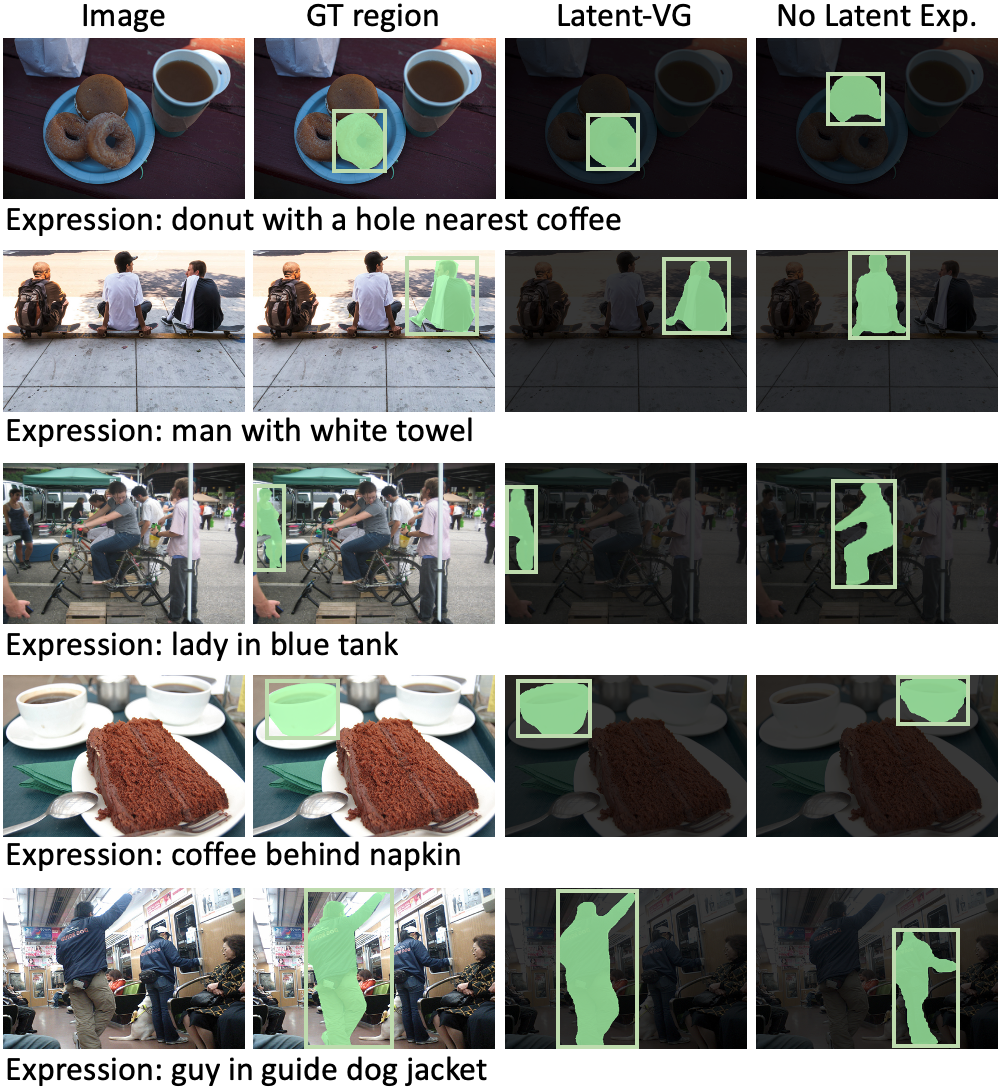}
    \caption{More qualitative analysis of Latent-VG compared to a model without any proposed methods (termed as \textit{No Latent Exp.}).} 
    \label{supfig:no_exp}
\end{figure}

In Fig.~\ref{supfig:no_exp}, we provide additional qualitative analysis of our Latent-VG compared to the base model without any of the proposed modules (termed as \textit{No Latent Exp.}).
The \textit{No Latent Exp.} model often fails to distinguish the target from other similar objects when the limited textual cues are given.
For instance, in the first row of Fig.~\ref{supfig:no_exp}, with the description ``\textit{donut with a hole nearest coffee}", the \textit{No Latent Exp.} model captures a non-targeted donut, while our approach precisely selects the targeted donut, indicating our superiority in capturing the target cues.

\subsection{IoU Scores for each Expression}
In Tab.~\ref{suptab:each_iou}, we report the IoU scores on the validation of RefCOCO+ for each expression, as well as the final prediction.
Each IoU score is calculated between the ground truth mask and the mask predicted by each expression.
As all expressions are optimized by the identical loss function, they exhibit similar performances.
However, the \textit{Latent Exps} consistently achieve slightly higher IoU scores than the \textit{Input Exp.}, even though they are derived from the input expression.
This demonstrates the enhanced ability of the latent expressions to deliver target details into the model.
\begin{table}[h]
    \centering
    \footnotesize
    \begin{tabular}{c|c@{\hspace{4pt}}c@{\hspace{4pt}}c|c}
    \hline
    Metric     & Input Exp. & Latent Exp.1 & Latent Exp.2 & Final Pred. \\ \hline
    mIoU     & 72.84 & \textbf{73.15} & \underline{73.07} & \textcolor{gray}{73.19} \\  \cdashline{1-5}[0.2pt/1pt]
    oIoU & 70.34 & \underline{70.60} & \textbf{70.63} &  \textcolor{gray}{70.68} \\
    \hline
    \end{tabular}
    \caption{IoU scores for each expression and the final prediction. Each IoU score is calculated between the ground truth mask and the mask predicted by each expression.
    }
    \label{suptab:each_iou}
\end{table}

\subsection{Convergence of each IoU Score.}
\begin{figure}[h]
    \centering
      \includegraphics[width=1\linewidth]{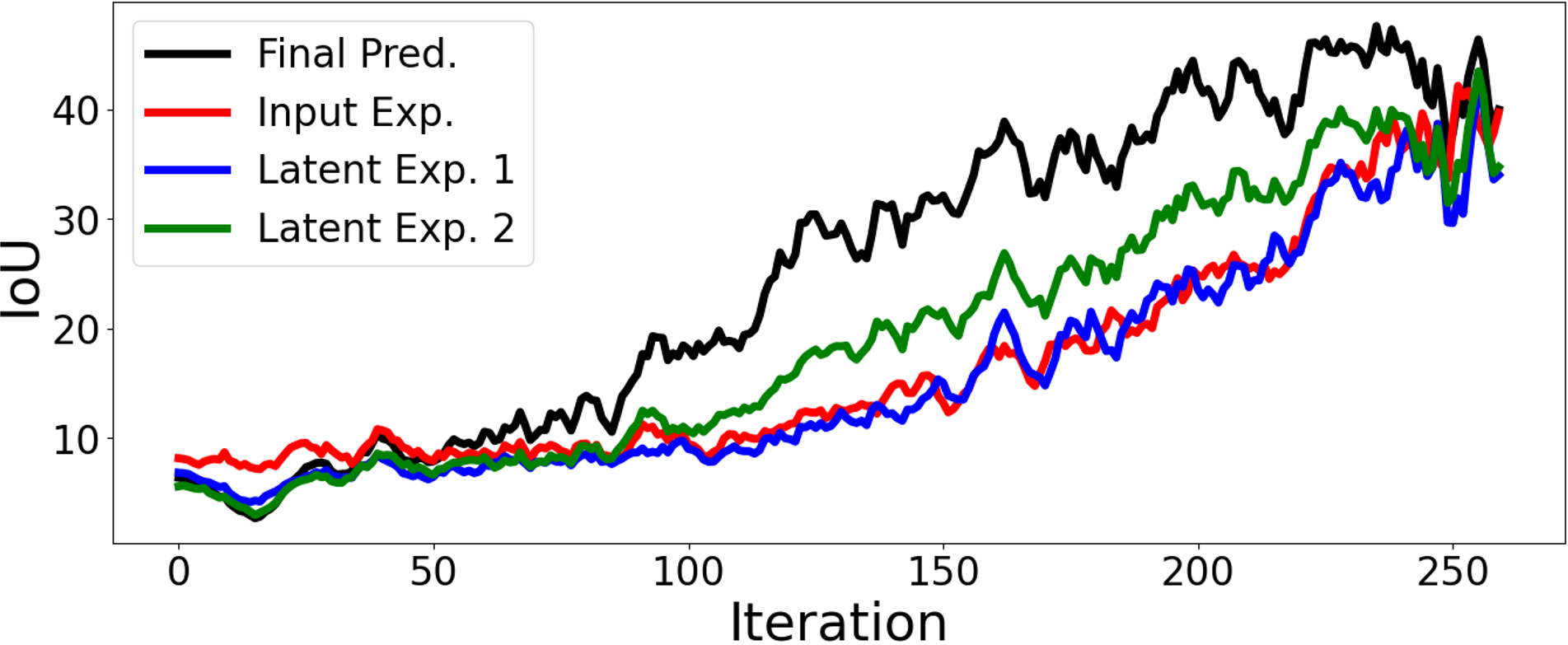}
    \caption{The convergence of IoU scores for each expression and the final prediction.} 
    \label{supfig:iou_conver} 
\end{figure}

Fig.~\ref{supfig:iou_conver} shows the convergence of IoU scores for each expression and the final prediction on the validation set of RefCOCO+.
In the early stages of training, individual expressions exhibit varied IoU scores; however, after approximately 250 iterations, all scores converge to similar values.
This convergence occurs because all expressions are optimized using the same learning objectives and are aligned similarly via the proposed contrastive loss.

\begin{figure*}[t]
    \centering
      \includegraphics[width=0.95\linewidth]{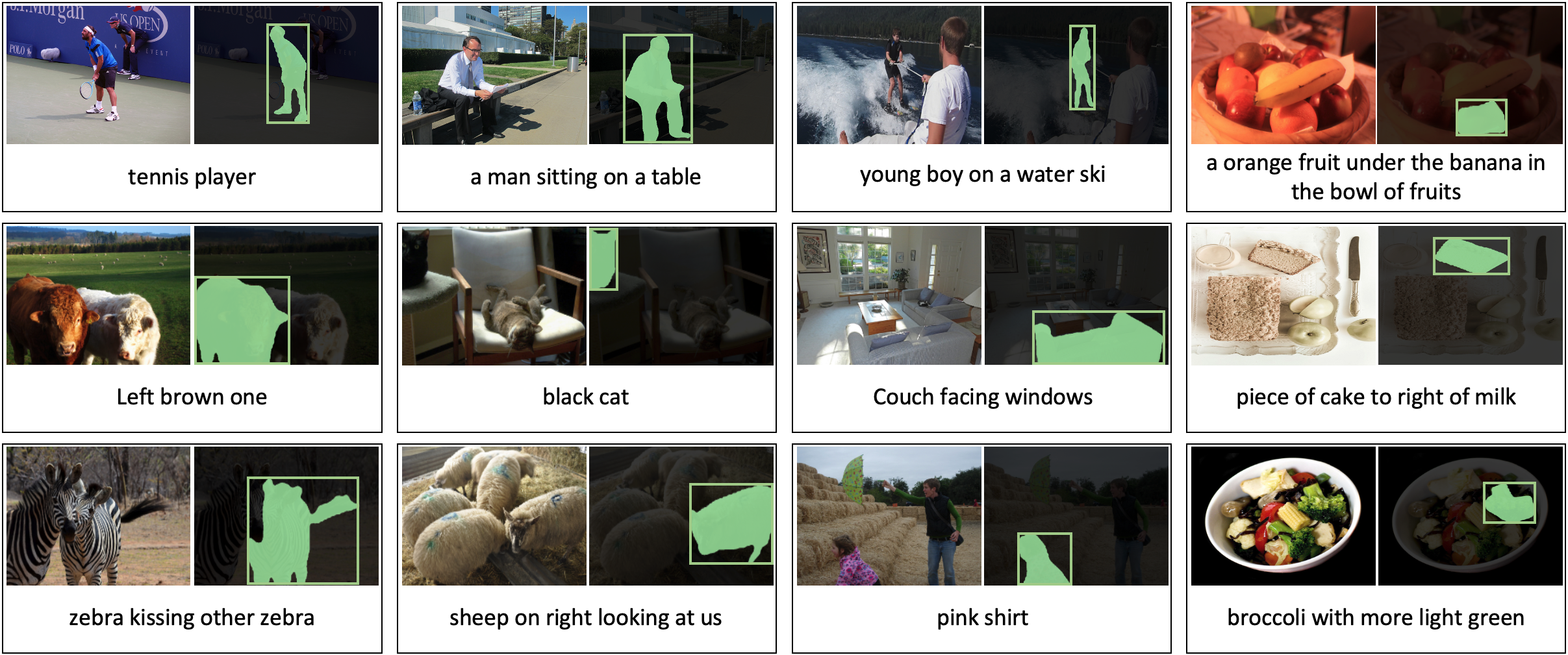}
      \vspace{-0.2cm}
    \caption{The visualization of segmentation and detection outputs of the proposed Latent-VG.} 
    \label{supfig:vis_ris_rec} 
    \vspace{-0.2cm}
\end{figure*}

\subsection{Examples of an Extracted Subject}
\label{exampls_subject}
In Sec 3.1 of the main manuscript, we extract a subject token from the input textual tokens by applying a linear layer followed by a Gumble-softmax operation and an argmax function.
The linear layer that generates subject logits is trained end-to-end using the framework learning objectives, without explicit supervision for the subject.
Tab.~\ref{tab:extracted_subject} presents examples of the extracted subject tokens.
In many cases, the correct subject is successfully extracted, probably because positioning the extracted token at the beginning of the latent expressions encourages accurate selection.
However, in some failure cases, the extracted token does not correspond exactly to the true subject but instead captures a crucial keyword distinguishing the target (\eg, for ``\textit{elephant in back}", the token ``\textit{back}" is selected).
Moreover, when the true subject consists of multiple words (\ie, mobile phone), only a single token (\eg, mobile) is extracted, which may not fully represent the subject.
We plan to explore these limitations in the future.
\begin{table}[t]
\centering{}
\footnotesize
\begin{tabular}{l}
\toprule
Correct Examples: \\
\midrule
\hspace{0.2cm} gray \colorbox{pink}{cat} \\
\hspace{0.2cm} bundle of \colorbox{pink}{broccoli} \\
\hspace{0.2cm} sugar powdered \colorbox{pink}{donut} \\
\hspace{0.2cm} \colorbox{pink}{sprinkle} even with face almost \\
\hspace{0.2cm} the \colorbox{pink}{zebra} on the left in the right hand picture \\
\hspace{0.2cm} a small \colorbox{pink}{girl} starring at something along with her elder sister \\
\hspace{0.2cm} a \colorbox{pink}{glass} with napkins and utensils inside of it sitting near a pizza \\
\toprule
Failure Examples: \\
\midrule
\hspace{0.2cm} \underline{elephant} in \colorbox{pink}{back} \\
\hspace{0.2cm} the \underline{\colorbox{pink}{man}'s hat} \\
\hspace{0.2cm} the \underline{vehicle} on the \colorbox{pink}{left} of the row \\
\hspace{0.2cm} the \underline{\colorbox{pink}{mobile} phone} with a number 2125 towards the top right side \\

\bottomrule
\end{tabular}
\vspace{-0.2cm}
\caption{Examples of the extracted \colorbox{pink}{subject} in the input sentence.}
\vspace{-0.2cm}
\label{tab:extracted_subject}
\end{table}




\vspace{-0.3cm}

\section{Further Discussion}
\label{supple:limitation}

\subsection{Limitations}
\begin{figure}[h]
    \centering
      \includegraphics[width=0.95\linewidth]{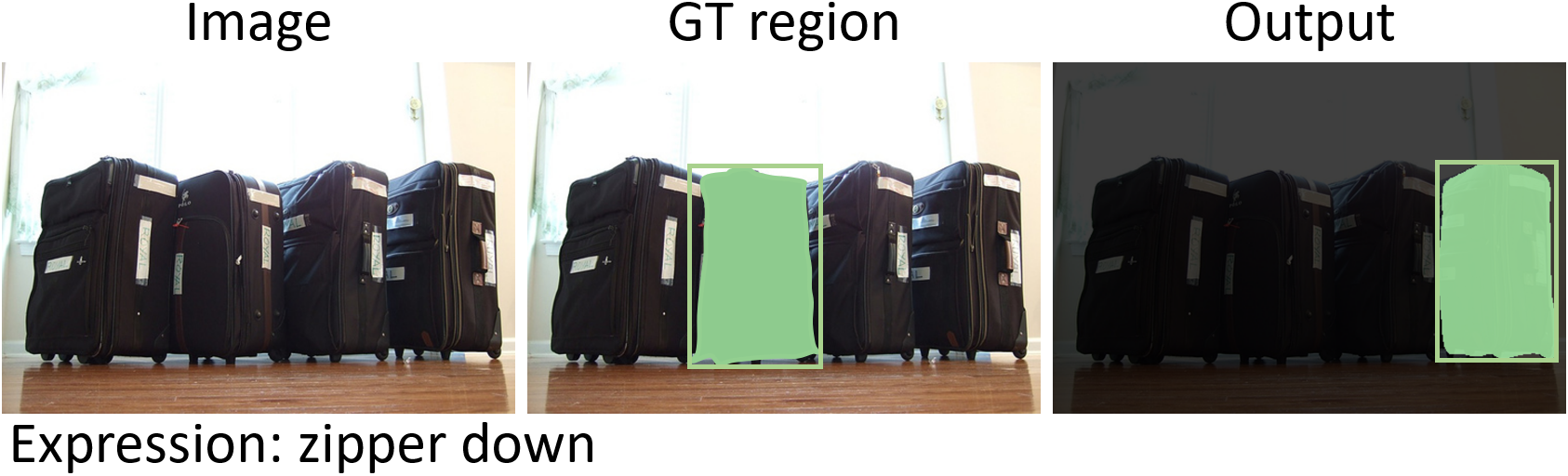}
      \vspace{-0.2cm}
    \caption{The failure example of our Latent-VG.} 
    \label{supfig:reliance on} 
    \vspace{-0.6cm}
\end{figure}

We discuss several limitations in our Latent-VG below:
\vspace{-0.4cm}
\paragraph{The Reliance on the Input Expression.}
Our latent expressions are initialized from the input textual tokens, leading to the inevitable dependence on the input textual semantics.
To mitigate this, we introduce to the dropout in Section 3.1 and select target-related patches as broad regions beyond the target area in the Visual Concept Injector from Sec. 3.2. in the main manuscript.
Despite these efforts, our framework remains sensitive to the semantics of the input text.
For example, as shown in Fig.~\ref{supfig:reliance on}, when identifying a suitcase with a downed zipper, the model must distinguish subtle differences in zipper locations among four suitcases.
These extremely limited input cues lead our model to an incorrect suitcase.
We tried to capture novel semantics outside of the input semantics in the latent representation, but if the input semantics occupy too small a portion of the target visual area, our model could select non-targeted objects.

\begin{figure}[h]
    \centering
      \includegraphics[width=0.95\linewidth]{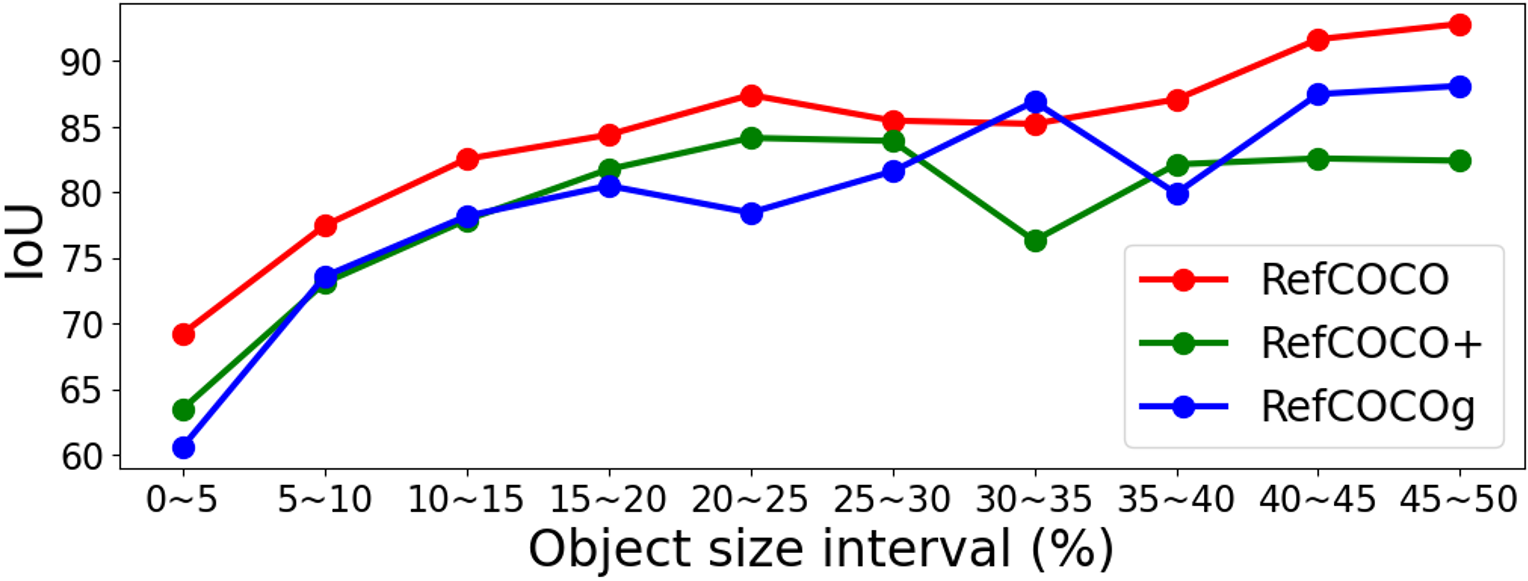}
      \vspace{-0.2cm}
    \caption{The IoU scores of our methods as the object size ratio.
    }
    \label{supfig:small_object} 
    \vspace{-0.2cm}
\end{figure}

\paragraph{The Weakness on the Small Size of Object.}
Since our method does not explicitly address the object scale variations, it performs less effectively on small-scale objects, as illustrated in Fig.~\ref{supfig:small_object}.
To assess this, we analyze the IoU scores based on the object size ratio (\ie target object size divided by total image size).
Our results reveal that performance drops significantly for object ratios in $0\%- 5\%$ and $5\%- 10\%$, showing a limitation in localizing small objects.
We plan to investigate this issue further in future work.

\begin{figure*}[t]
    \centering
      \includegraphics[width=0.9\linewidth]{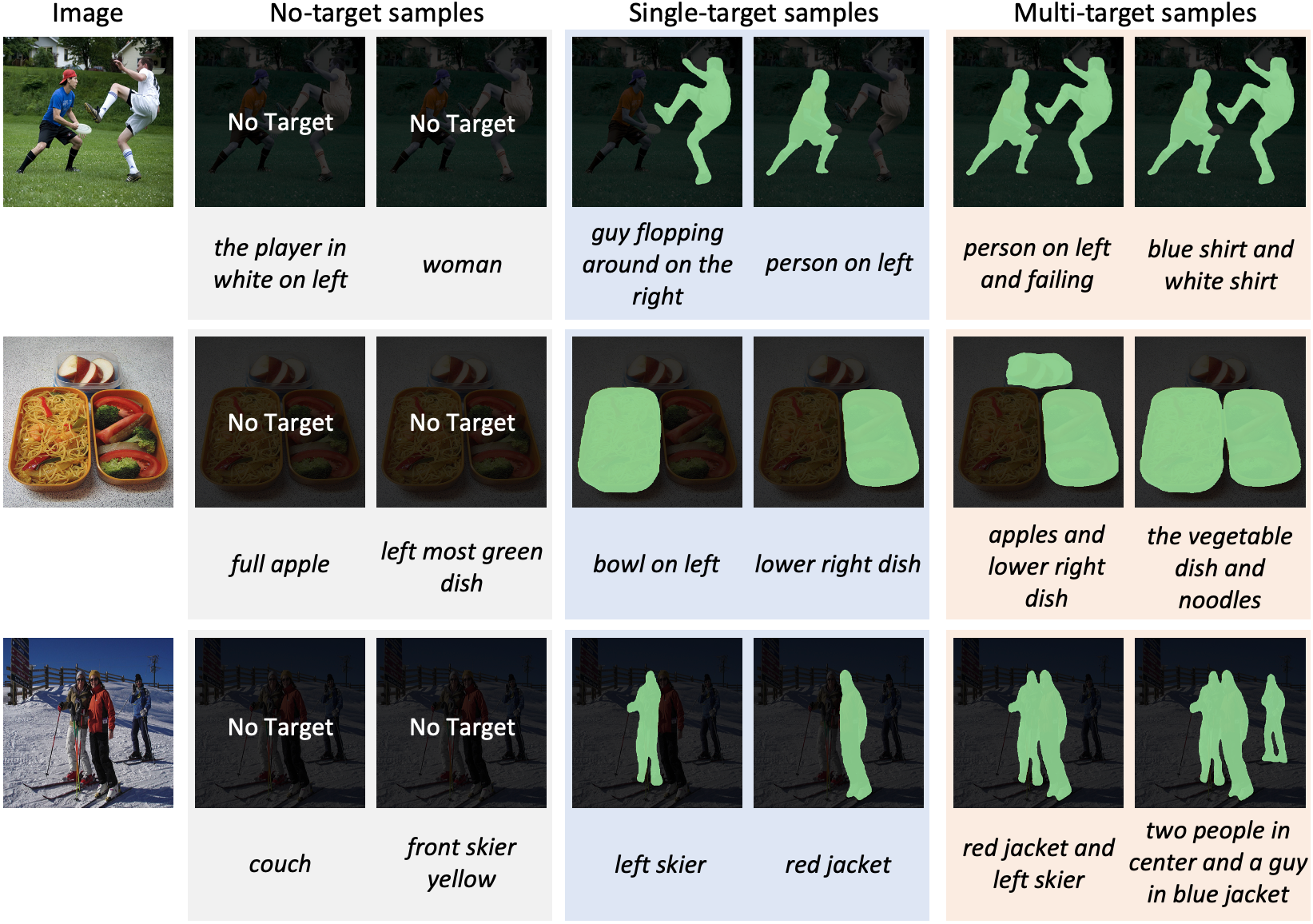}
      \vspace{-0.2cm}
    \caption{The visualization of predicted results of our Latent-VG for the GRES task.} 
    \label{supfig:vis_gres} 
    \vspace{-0.2cm}
\end{figure*}

\paragraph{The Incorrect Subject Selection.}
As discussed in Sec.~\ref{exampls_subject} and shown by the failure examples in Tab.~\ref{tab:extracted_subject}, our design of subject selection can occasionally extract an incorrect subject.
Although the incorrectly chosen token may sometimes capture a key distinguishing word (\eg, ``\textit{left}" in ``\textit{the vehicle on the left of the row}", as reported in Tab.~\ref{tab:extracted_subject}), this is not our intended outcome.
We will explore alternative strategies for precise subject extraction, such as representing a subject token by operating a weighted sum over all textual tokens or incorporating subject supervision obtained via natural language processing (NLP) tools.


Despite these limitations, our methods achieve state-of-the-art performances in RIS, REC, and GRES tasks, demonstrating the effectiveness of our approach in leveraging latent expressions to capture novel semantics outside the input text.



%



\subsection{Social Impact}
Our work may inadvertently propagate biases present in the training data, leading to unintended ethical concerns.
In addition, the capability to generate highly specified segmentation by users could be exploited for misinformation or deceptive media manipulation, further highlighting the importance of careful monitoring and regulation.

\section{Visualizations}
\label{supple:more_visualization}

\subsection{Visualization on RIS and REC}
In Fig.~\ref{supfig:vis_ris_rec}, we present the visualizations of the outputs generated by our Latent-VG for the referring image segmentation (RIS) and referring expression comprehension (REC) on the RefCOCO(+/g) datasets.

\subsection{Visualization on GRES}
In Fig.~\ref{supfig:vis_gres}, we visualize the examples of prediction results of our Latent-VG on the GRefCOCO dataset.
Our framework demonstrates a strong ability to handle different referring descriptions, and the no-, single-, and multi-target scenarios for the same input images.



\end{document}